\documentclass{article}
\usepackage{iclr2027_conference,times}

\usepackage{amsmath,amsfonts,bm}

\def\eqref#1{equation~\ref{#1}}
\def\1{\bm{1}}

\DeclareMathAlphabet{\mathsfit}{\encodingdefault}{\sfdefault}{m}{sl}
\SetMathAlphabet{\mathsfit}{bold}{\encodingdefault}{\sfdefault}{bx}{n}

\usepackage[hidelinks]{hyperref}
\usepackage{url}
\usepackage{graphicx}
\usepackage{wrapfig}
\usepackage{needspace}
\usepackage{booktabs}
\usepackage{amsmath}
\usepackage{amssymb}
\usepackage{xcolor}
\usepackage{colortbl}
\usepackage{array}
\usepackage{tabularx}
\usepackage{multirow}
\usepackage{enumitem}
\usepackage{caption}
\usepackage{placeins}
\usepackage{comment}
\iclrfinalcopy 
\definecolor{dropcyan}{RGB}{0, 103, 122}
\definecolor{tableblue}{RGB}{220,230,238}
\definecolor{tablebluedark}{RGB}{80,105,125}

\title{Representation Dynamics Reveal Semantic Saliency and Similarity for Visual Token Pruning in MLLMs}

\author{%
\textbf{Weixuan Li\textsuperscript{1}}\quad
\textbf{Zikun Zhou\textsuperscript{1}}\quad
\textbf{Xinyi Zhuang\textsuperscript{1}}\quad
\textbf{Xinyan Guo\textsuperscript{1,2}}\quad
\textbf{Rui Tian\textsuperscript{1,2}}\\
\textbf{Chuyao Zhang\textsuperscript{1}}\quad
\textbf{Lin Gao\textsuperscript{1}}\\[6pt]
\textsuperscript{1}Harbin Institute of Technology, Shenzhen\qquad\textsuperscript{2}Shenzhen Loop Area Institute\\[3pt]
{\normalfont\texttt{lwx20030918@gmail.com}}%
}

\newcommand{\ours}{MSDG-Prune}

\begin{document}

\maketitle
\suppressfloats 

\begin{abstract}
Multimodal large language models (MLLMs) incur high inference latency from long visual token sequences. Existing pruning methods commonly use attention maps or output features to estimate token importance or redundancy. Several recent approaches also exploit representation changes, but when and how these changes reflect foreground saliency and semantic consistency remain insufficiently understood.
We analyze visual token representation dynamics across encoder depth and uncover two findings.
First, the relationship between token update magnitudes and foreground saliency is layer-dependent: large token updates concentrate on foreground regions in two depth intervals, separated by several sink-dominated layers at intermediate depths.
Second, similarities between token update directions better distinguish same-class from different-class tokens than those between encoder output features.
Building on these findings, we propose \ours{}, a training-free method that uses update magnitudes and directions to preserve salient and diverse visual information. Specifically, we group tokens by update-direction similarity and use query-weighted saliency derived from update magnitudes across a chosen depth window for group-wise token pruning.
Extensive experiments across four MLLMs demonstrate the effectiveness and generalizability of \ours{}. On LLaVA-NeXT, it retains 91.9\% of uncompressed performance on average with only 5.6\% of visual tokens, while achieving a 7.8$\times$ prefilling speedup. Code is available at \url{https://github.com/liweixuan-hitsz/MSDG-Prune}.

\end{abstract}

\section{Introduction}
\label{sec:intro}

Multimodal Large Language Models (MLLMs) have demonstrated strong capabilities in visual understanding and reasoning~\citep{achiam2023gpt,li2025mini}.
These models connect vision encoders to pretrained LLMs and represent each image with hundreds to thousands of visual tokens~\citep{liu2023visual,bai2025qwen25vltechnicalreport}.
Such long sequences increase prefilling latency and KV-cache memory usage, yet contain substantial redundancy.
Prior work shows that many visual tokens can be removed with little performance degradation~\citep{bolya2022token,shang2025llava}, making pruning a practical approach to efficient inference.
The key challenge is to remove redundant tokens while preserving visual evidence needed to answer the query.

\begin{figure}[t]
\centering
\captionsetup{skip=4pt,belowskip=0pt}
\includegraphics[width=\linewidth]{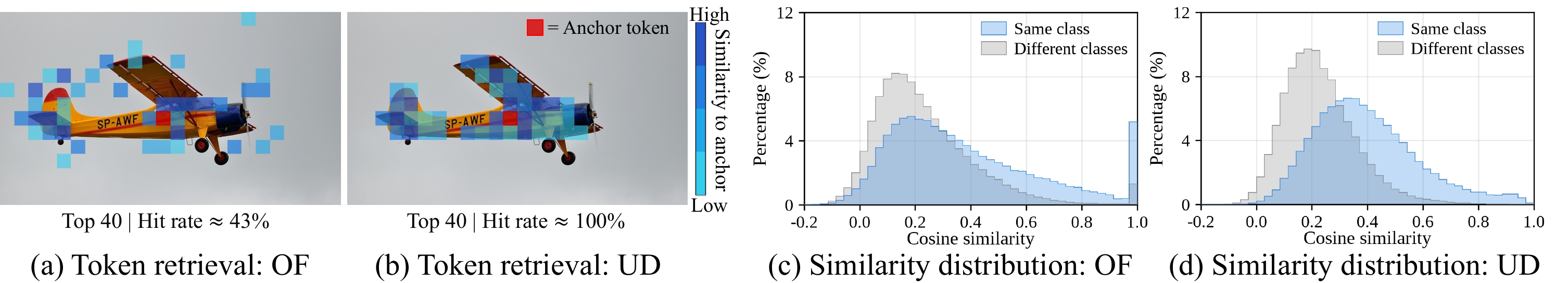}
\caption{
\textbf{Output features (OF) versus update directions (UD) for semantic similarity in LLaVA-1.5.} Panels (a) and (b) visualize the top 40 tokens ranked by cosine similarity to the airplane anchor token (\textcolor{red}{red}), using OF and UD, respectively. Panels (c) and (d) show cosine similarity distributions on COCO 2017 computed using OF and UD, respectively. For each anchor token, we compute cosine similarities to other tokens and group the resulting pairs by whether they share the same semantic class as the anchor token. \textcolor[RGB]{55,138,226}{Blue} and \textcolor[RGB]{142,142,142}{gray} curves represent same-class and different-class pairs, respectively. The horizontal axis indicates cosine similarity, and the vertical axis shows the percentage of pairs within each similarity bin. Compared with OF, UD shows less overlap between the two groups, indicating better semantic discrimination.}
\label{fig:score-distributions}
\end{figure}

Two widely studied approaches to visual token pruning are \textit{attention-based} and \textit{diversity-based} methods.
\textit{Attention-based} methods~\citep{chen2024image,yang2025visionzip,liu2026hiprune} mainly derive visual saliency from attention maps inside the vision encoder or LLM.
However, final-layer encoder attention can overemphasize high-norm outlier tokens with limited information~\citep{darcet2024vision,Fan_2026_CVPR,takezoe2026learnpruner}, while text-to-vision attention in LLMs can exhibit positional bias~\citep{endo2025feather}. These limitations make attention scores less reliable for token pruning.
\textit{Diversity-based} methods estimate token redundancy from cosine similarity between token representations, using high feature similarity as evidence of overlapping visual information~\citep{alvar2025divprune,wen2025stop}.
Nevertheless, cosine similarity between output features of the vision encoder does not align well with semantic consistency, evidenced by the overlap between the distributions in Figure~\ref{fig:score-distributions}(c).
Figure~\ref{fig:score-distributions}(a) illustrates an example: only a few retrieved tokens (blue) belong to the same class as the anchor token (red).
Hence, using this similarity to infer redundancy may lead to the removal of semantically distinct visual tokens needed to answer the query.

\begin{figure*}[t]
\centering
\includegraphics[width=\textwidth]{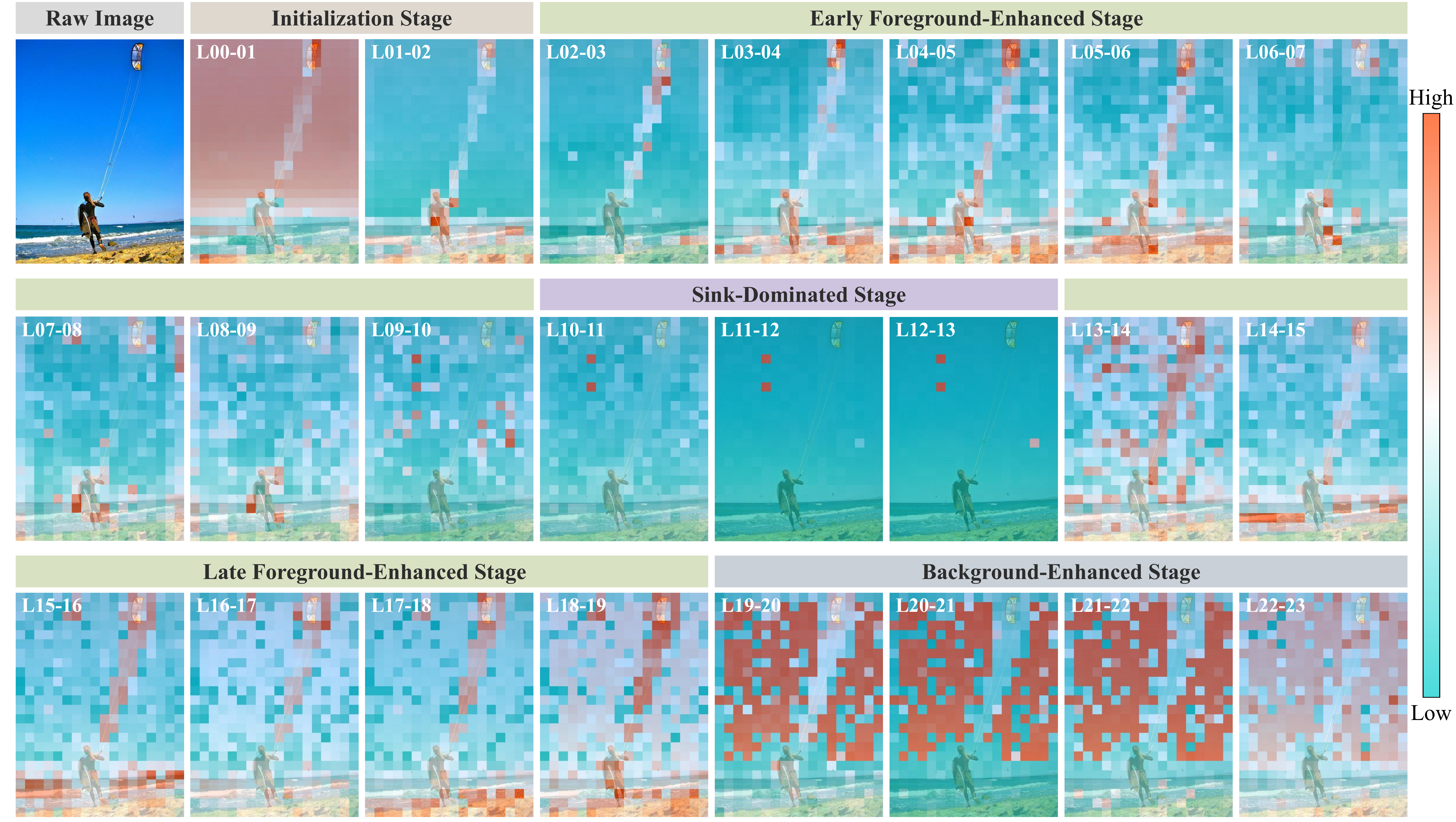}
\caption{
\textbf{Layer-wise token update magnitudes in the vision encoder of LLaVA-1.5.}
Each map overlays patch-token update magnitudes across a Transformer block on the input image. L00-01 denotes the change from token states \textbf{entering} Layer 0 to those \textbf{entering} Layer 1. Because LLaVA-1.5 uses the penultimate-layer output of its CLIP ViT-L/14@336~\citep{radford2021learning,liu2024improved} vision encoder as input to the visual projector, the visualization ends at L22-23 and omits L23-24. The maps illustrate depth-dependent spatial patterns in token updates, including foreground enhancement before and after the sink-dominated stage.}
\label{fig:layer-updates-intro}
\end{figure*}

\setlength{\textfloatsep}{8pt plus 2pt minus 4pt}
\AddToHookNext{cmd/@cflt/before}{\setlength{\textfloatsep}{11pt}}
Beyond the static attention maps and token features, could the way token representations change across layers provide more reliable evidence for pruning? Recent studies have explored representation changes as pruning cues~\citep{choi2025representation,li2026transprune}. However, understanding when these changes reflect semantic saliency and similarity warrants closer examination.
To this end, we analyze the magnitude and direction of visual token updates within the encoder.
Our analysis reveals a stage-dependent relationship between update magnitude and foreground saliency, as shown in Figure~\ref{fig:layer-updates-intro}.
Foreground enhancement occurs in both early and late stages, interrupted by large updates concentrated on isolated sink tokens, while background responses become more prominent near the encoder output.
Thus, the usefulness of update magnitude as a cue for foreground saliency depends on encoder depth. This finding motivates estimating visual saliency from representation changes within a selected foreground-enhanced window.

Beyond update magnitude, we examine whether changes in representation direction reveal semantic relationships among tokens, given the mismatch between output-feature similarity and semantic consistency.
We find that similarity between update directions better distinguishes same-class from different-class token pairs than output-feature similarity, as shown in Figure~\ref{fig:score-distributions}(c) and (d).
This finding motivates using update-direction similarity to form semantically coherent groups, providing a basis for preserving distinct visual information during pruning. Both findings are further supported by observations across multiple MLLMs (Appendices~\ref{app:feature-dynamics} and~\ref{app:direction-analysis}).

Motivated by these findings, we propose \textbf{\ours{}}, a training-free visual token pruning method that leverages update \textbf{M}agnitudes for \textbf{S}aliency estimation and update \textbf{D}irections for token \textbf{G}rouping. Specifically, we estimate saliency within a selected foreground-enhanced window and weight it by query relevance. These query-weighted saliency scores guide budget allocation across the semantic groups and token selection within each group. This design preserves visual diversity while prioritizing salient, query-relevant tokens.
Our method prunes visual tokens before LLM prefilling without extracting attention maps, reducing prefilling computation and KV-cache memory while remaining compatible with FlashAttention~\citep{dao2022flashattention,dao2024flashattention}. Extensive evaluations across four MLLMs show that \ours{} achieves favorable performance against state-of-the-art methods. At 11.1\% token retention, it outperforms PruneSID~\citep{fang2026prune} by 2.1 and 1.5 percentage points on Mini-Gemini and Qwen2.5-VL, respectively. On LLaVA-NeXT, it retains 91.9\% relative performance with only 5.6\% of visual tokens and achieves a 7.8$\times$ prefilling speedup over the uncompressed model.
Our contributions are:
\begin{itemize}[topsep=3pt,partopsep=0pt,itemsep=1pt,parsep=0pt]
    \item We analyze visual token representation dynamics, revealing a depth-dependent relationship between update magnitude and foreground saliency and better semantic discrimination from update directions than output features.
    
    \item Building on these findings, we propose \ours{}, a training-free visual token pruning method that uses update magnitudes and directions to more effectively preserve diverse and salient visual information.
    
    \item Extensive quantitative and qualitative evaluations across four MLLMs demonstrate the effectiveness and generalizability of \ours{}. On LLaVA-1.5, it retains 95.4\% of uncompressed performance on average at 11.1\% token retention.
\end{itemize}

\begingroup
\makeatletter
\renewcommand{\section}{\@startsection{section}{1}{\z@}{-2.0ex}{1.5ex}{\large\sc\raggedright}}
\section{Related Work}
\label{sec:related}
\makeatother
\endgroup
We review visual token pruning methods according to their main pruning cues, including attention, representation similarity, and changes in token representations.

Attention-based methods use attention scores to guide token selection.
FastV~\citep{chen2024image} and SparseVLM~\citep{zhang2025sparsevlm} use attention within the LLM to identify visual tokens for pruning.
VisionZip~\citep{yang2025visionzip} and LLaVA-PruMerge~\citep{shang2025llava} combine token selection based on vision encoder attention with similarity-based merging.
HiPrune~\citep{liu2026hiprune} uses attention patterns at different encoder depths to retain complementary token types.

Similarity-based methods use relationships between token representations to reduce redundancy.
ToMe~\citep{bolya2022token} merges similar tokens within the encoder, while DivPrune~\citep{alvar2025divprune} selects tokens by maximizing feature diversity.
DART~\citep{wen2025stop} uses similarity to visual and textual pivot tokens to identify redundant visual tokens within the LLM.
CDPruner~\citep{NEURIPS2025_2433fec2} combines pairwise feature similarity with instruction relevance in a determinantal point process kernel.
PruneSID~\citep{fang2026prune} forms groups through principal semantic component analysis, suppresses redundant tokens within each group, and allocates retention quotas according to the remaining group sizes.

Recent methods also use representation changes as pruning cues.
Representation Shift~\citep{choi2025representation} scores tokens from differences between the inputs and outputs of network operations, without extracting attention maps.
TransPrune~\citep{li2026transprune} combines module-level magnitude and angular changes with instruction-guided attention for progressive pruning within the LLM.
EvoCut~\citep{lu2026evocut} clusters update directions between adjacent encoder layers and accumulates deviations from cluster centers to rank tokens globally.
These methods primarily use representation changes to construct token-level importance scores.

In \ours{}, visual saliency is estimated from the update magnitude over a selected window in the late foreground-enhanced stage.
Semantic groups are formed using directions of change between normalized early and late encoder representations.
These groups serve as units for budget allocation and intra-group selection, both guided by visual saliency and query relevance computed from embedding similarity.
The resulting selection is performed before LLM prefilling without extracting attention maps.
\section{Representation Dynamics of Visual Tokens}
\label{sec:representation-dynamics}
In this section, we analyze representation dynamics within the vision encoder to inform the design of visual token pruning methods. Specifically, we examine the magnitude and direction of visual token updates across encoder depth. Herein, we consider a vision encoder with $L$ layers, indexed from $0$ to $L-1$.
Each layer $\ell$ updates the representation of visual token $i$ from $h_i^\ell$ to $h_i^{\ell+1}$, where $h_i^\ell \in \mathbb{R}^{d_v}$ and $d_v$ is the hidden dimension of the encoder.

\subsection{Analysis of Token Update Magnitudes}
\label{sec:update-magnitude-foreground-saliency}
\begin{wrapfigure}{R}{0.45\textwidth}
    \vspace{-5mm}
    \centering
\includegraphics[width=\linewidth]{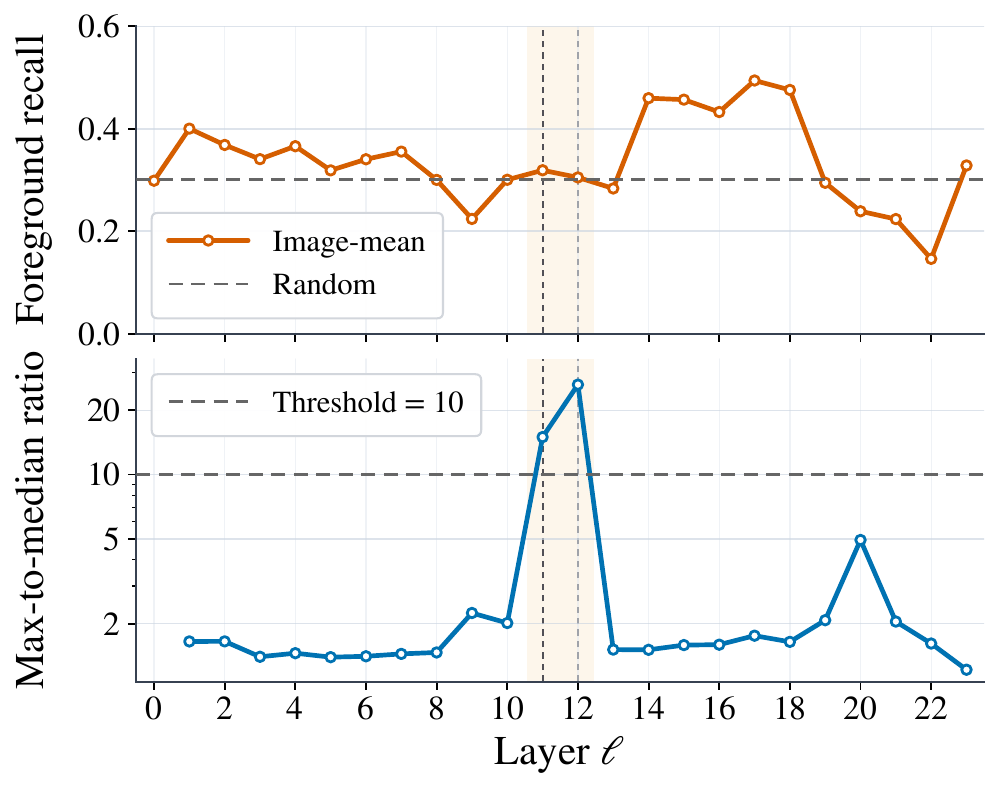}
\captionsetup{font=small,skip=4pt}
\caption{Layer-wise analysis of token update magnitudes in LLaVA-1.5 on COCO 2017 validation images.
Top: foreground recall of the top $30\%$ of tokens ranked by update magnitude, with random selection as the dashed baseline.
Bottom: the max-to-median update magnitude ratio, $\max_i v_i^\ell / \operatorname{median}_i v_i^\ell$, with a dashed diagnostic threshold of $10$.
Both metrics are computed per image and then averaged across images.
The highlighted Layers 11--12 exhibit sharp ratio peaks, marking the sink-dominated stage in the corresponding  vision encoder.}
\label{fig:sink-ratio-llava}
\vspace{-6mm}
\end{wrapfigure}
To gain insight into how visual token representations evolve across encoder depth, we begin with a qualitative examination of their update magnitudes.
For each layer $\ell$, we map the update magnitudes $v_i^\ell=\lVert h_i^{\ell+1}-h_i^\ell\rVert$ to the corresponding image regions, as illustrated in Figure~\ref{fig:layer-updates-intro}. The spatial distribution of token update magnitudes exhibits distinct stage-wise patterns across encoder depth, which we broadly characterize as five successive stages: 

\noindent\textbf{Initialization stage.} 
In the first few layers, token updates appear to primarily reflect low-level visual properties, such as local smooth or textured regions.

\noindent\textbf{Early foreground-enhanced stage.} 
Relatively large updates become more apparent over foreground regions in the early encoder layers. This tendency emerges gradually, with no sharp boundary separating this stage from initialization.

\noindent\textbf{Sink-dominated stage.} At intermediate depths, large updates concentrate on a few spatially isolated \emph{sink tokens}, which are high-norm patch tokens and often located in background regions. Prior work suggests that sink tokens are used for internal computation and storing global information~\citep{darcet2024vision}, but carry little image-specific semantic information in CLIP~\citep{Fan_2026_CVPR}.
Strong sink responses do not necessarily indicate visual saliency.

\noindent\textbf{Late foreground-enhanced stage.}
After the sink-dominated stage, relatively large updates again extend over foreground regions. This spatially extended pattern can be observed across several layers, unlike the large updates concentrated on a few isolated sink tokens.

\noindent\textbf{Background-enhanced stage.}
Near the encoder output, large updates increasingly occur in background regions. This shift suggests that deeper updates are not necessarily better saliency cues.

Building on these visualizations, we quantitatively examine the depth-dependent patterns on COCO 2017~\citep{lin2014microsoft}. For each image and layer, we select the top $30\%$ of tokens by update magnitudes and measure foreground recall against COCO instance annotations. This metric measures how much annotated foreground area is covered by the selected tokens. We also compute the concentration ratio $\max_i v_i^\ell/\operatorname{median}_i v_i^\ell$ to quantify how strongly the largest update exceeds the typical token update.
Both statistics are averaged across images and plotted against encoder depth in Figure~\ref{fig:sink-ratio-llava}.

The concentration ratio exhibits pronounced peaks at the middle layers, consistent with the sink-dominated patterns across all examined models.
These peaks are not accompanied by increased foreground recall, indicating that exceptionally large updates do not necessarily correspond to foreground regions.
In contrast, foreground recall is elevated in both early and late foreground-enhanced stages before declining near the encoder output.
This depth dependence motivates using update magnitudes from an early or late foreground-enhanced stage as a training-free cue for foreground saliency.

\subsection{Analysis of Token Update Directions}
\label{sec:update-direction-semantic-consistency}
We next turn to changes in representation direction across encoder depth and investigate whether tokens undergoing similar direction changes are semantically related. We compare the representation of each token at an early reference state after the initialization stage, $h_i^{\ell_0}$, with its late encoder state, $h_i^{\ell_d}$, where $\ell_0=2$ and $\ell_d=L-1$.
To examine direction changes independently of the norms of the early and late representations, we normalize both representations to unit length. We define the \emph{update direction} as the unit direction of the displacement between these normalized states:
\begin{equation}
    \Delta h_i
    = \frac{h_i^{\ell_d}}{\lVert h_i^{\ell_d}\rVert}
    - \frac{h_i^{\ell_0}}{\lVert h_i^{\ell_0}\rVert},
    \qquad
    d_i = \frac{\Delta h_i}{\lVert\Delta h_i\rVert}.
\label{eq:delta-h}
\end{equation}
We use cosine similarity $d_i^\top d_j$ to measure the alignment between token update directions and compare it with the commonly used output-feature similarity.
We first qualitatively compare the spatial patterns of the two similarity measures on LLaVA-1.5, as shown in Figure~\ref{fig:score-distributions}(a) and (b). For the airplane anchor token (marked in red), output-feature similarity assigns high scores to tokens both on and outside the airplane, whereas update-direction similarity better distinguishes airplane tokens from surrounding tokens.

To examine whether this observation holds across images, we analyze token-pair similarities on COCO 2017 using COCO-Stuff semantic annotations~\citep{caesar2018coco}.
For each anchor token, we compute cosine similarities to other labeled tokens in the same image and separate the pairs according to whether they share the semantic class with the anchor.
We aggregate these scores across anchors to obtain same-class and different-class similarity distributions for output features and update directions.
As shown in Figure~\ref{fig:score-distributions}(c) and (d), update-direction similarity yields less overlap between the two distributions, indicating better semantic discrimination. This finding supports update-direction similarity as a cue for semantically coherent token grouping.

Additional visualizations and quantitative results across models further support our findings on update magnitude and direction (Appendices~\ref{app:feature-dynamics} and~\ref{app:direction-analysis}).

\section{\ours{}}
\label{sec:method}

\subsection{Problem Formulation and Method Overview}
\label{sec:prelim}

\noindent\textbf{Problem formulation.}
The input image $I$ of the MLLM is generally encoded and projected into a sequence of visual tokens $\bm{X}=[\bm{x}_1,\ldots,\bm{x}_N]\in\mathbb{R}^{N\times D}$, where $N$ is the number of visual tokens before pruning and $D$ is the LLM embedding dimension.
Given a retention budget $B\ll N$, the goal of visual token pruning is to select an index set $\mathcal S\subseteq\{1,\ldots,N\}$ with $|\mathcal S|=B$ before LLM prefilling while preserving important and diverse visual information.
The retained tokens form $\widetilde{\bm{X}}$ in their original order and are passed to the pretrained LLM together with the tokenized query.

\begin{figure}[t]
\centering
\includegraphics[width=\linewidth]{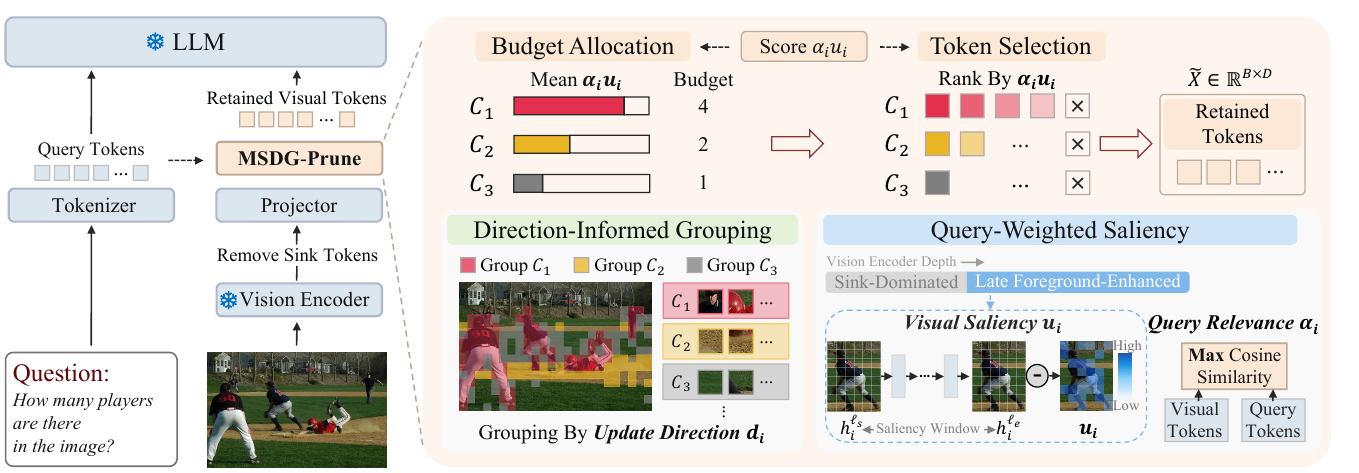}
\caption{Overview of \ours{}.
After sink filtering, update magnitudes over a selected foreground-enhanced window provide visual saliency scores, while update-direction similarity organizes tokens into semantic groups.
Visual saliency and query relevance jointly guide budget allocation across groups and token selection within each group.}
\label{fig:pipeline}
\end{figure}

\noindent\textbf{Method overview.}
Figure~\ref{fig:pipeline} summarizes the training-free pruning pipeline of \ours{}. 
We first remove sink tokens using a threshold on a designated encoder coordinate (Appendix~\ref{app:sink-detection}) and denote the remaining set by $\mathcal V$.
For each candidate token in $\mathcal V$, \ours{} leverages the query-weighted saliency to assess its importance for the MLLM to respond to the query.
Furthermore, \ours{} groups tokens by update-direction similarity and performs pruning within each group to preserve diversity among the retained tokens.
Next, we detail these two core methodology designs.

\subsection{Query-weighted Saliency Estimation}
\label{sec:foreground-window-saliency-estimation}
We construct a query-weighted saliency score to assess candidate tokens by combining their visual saliency with relevance to the textual query.
Our analysis in Section~\ref{sec:update-magnitude-foreground-saliency} motivates estimating visual saliency from representation changes within the foreground-enhanced stage.
For each candidate token $i\in\mathcal V$, we define $u_i = \lVert h_i^{\ell_e} - h_i^{\ell_s} \rVert,$ where $\ell_s$ and $\ell_e$ delimit the selected layer window.

We leave a one-layer buffer after the sink-dominated stage and use a window of five consecutive layer updates as an empirical default.
Ablations on window placement and width support this choice (Appendices~\ref{app:saliency-window}
and~\ref{app:saliency-window-length}).

The resulting saliency score is independent of the textual query. To account for the relevance of each visual token to the query, we weight its visual saliency score by the maximum cosine similarity between its projected embedding and the query token embeddings.
Let $\mathcal Q$ denote the set of query token indices, and let $\bm{t}_j$ denote the input embedding of query token $j$.
For each projected visual token $\bm{x}_i$, we compute its query relevance score as
\begin{equation}
    \alpha_i =
    \max_{j\in\mathcal Q}
    \frac{\bm{x}_i^\top\bm{t}_j}
    {\lVert\bm{x}_i\rVert\lVert\bm{t}_j\rVert}.
    \label{eq:query-similarity}
\end{equation}
The query-weighted saliency score is then defined as $s_i=\alpha_i u_i$.

\subsection{Direction-Informed Group-Wise Pruning}
\label{sec:group-wise-pruning-draft}
To preserve diverse visual information, we adopt the group-wise pruning mechanism, where the tokens are grouped by update-direction similarity.
We then use the above query-weighted saliency to guide budget allocation across groups and select important tokens within each group.

Specifically, we partition $\mathcal V$ into $K$ groups using spherical $k$-means~\citep{hornik2012spherical} on the update directions $d_i$ defined in Equation~\ref{eq:delta-h}.
The grouping objective maximizes the sum of cosine similarities between token update directions and their assigned unit centroids:
\begin{equation}
    \max_{\{\mathcal C_c\},\{\mu_c\}}
    \sum_{c=1}^{K}\sum_{i\in\mathcal C_c}d_i^\top\mu_c
    \quad
    \text{s.t.}\ \lVert\mu_c\rVert=1,\quad c=1,\ldots,K,
    \label{eq:group-wise-clustering-draft}
\end{equation}
where $\{\mathcal C_c\}_{c=1}^{K}$ partitions $\mathcal V$ into $K$ nonempty groups, and $\mu_c$ is the unit centroid of group $\mathcal C_c$.
We initialize the centroids using deterministic farthest-point sampling and stop the iterations when a convergence criterion is met or a preset iteration limit is reached.
For each group, we compute its mean query-weighted saliency score $g_c$ and apply softmax across groups to obtain the target budget proportion $p_c$:
\begin{equation}
    g_c=\frac{1}{|\mathcal C_c|}\sum_{i\in\mathcal C_c}s_i,
    \qquad
    p_c=\frac{\exp(g_c)}{\sum_{r=1}^{K}\exp(g_r)}.
    \label{eq:group-wise-budget-draft}
\end{equation}

For $B\leq|\mathcal V|$, we allocate integer budgets $b_c$ to groups according to $p_c$, subject to $0\leq b_c\leq|\mathcal C_c|$ and $\sum_{c=1}^{K}b_c=B$.
Within each group $\mathcal C_c$, we retain the $b_c$ tokens with the highest query-weighted saliency scores $s_i$ and combine their indices to form the retained set $\mathcal S$.
Finally, we arrange the selected tokens in their original order to form $\widetilde{\bm{X}}$ and pass them to the LLM with the tokenized query.
Algorithm~\ref{alg:algorithm} summarizes the complete token selection procedure.
\section{Experiments}
\label{sec:exp}

Following the experimental protocol of PruneSID~\citep{fang2026prune}, we evaluate \ours{} on LLaVA-1.5~\citep{liu2024improved}, LLaVA-NeXT~\citep{liu2024llavanext}, Mini-Gemini~\citep{li2025mini} and Qwen2.5-VL~\citep{bai2025qwen25vltechnicalreport}.
These models respectively cover fixed-resolution encoding, image tiling, a dual-encoder architecture, and native dynamic resolution, allowing us to examine the effectiveness of representation dynamics across different visual processing pipelines.

We compare our method with FastV~\citep{chen2024image}, SparseVLM~\citep{zhang2025sparsevlm}, DART~\citep{wen2025stop}, DivPrune~\citep{alvar2025divprune}, VisionZip~\citep{yang2025visionzip}, HiPrune~\citep{liu2026hiprune}, EvoCut~\citep{lu2026evocut}, and PruneSID~\citep{fang2026prune}.
Evaluation is conducted using LMMs-Eval~\citep{zhang2025lmms} on benchmarks covering visual question answering, visual perception and reasoning, object hallucination assessment, text understanding, and high-resolution image understanding.
RelAcc.\ is the mean of per-benchmark scores normalized by the corresponding uncompressed scores, expressed as a percentage, with missing entries excluded.
Implementation settings and benchmark descriptions are provided in Appendices~\ref{app:exp-details} and~\ref{app:benchmarks}, respectively.

\begin{table}[hb]
\centering
\captionsetup{font=small,skip=4pt}
\caption{Performance comparison on LLaVA-1.5-7B.
\textit{Vanilla} refers to the uncompressed baseline.
The subscript after each method gives the publication venue.
}
\label{tab:llava15-main}
\begingroup
\fontsize{9}{11}\selectfont
\setlength{\tabcolsep}{2.5pt}
\renewcommand{\arraystretch}{0.9}
\begin{tabularx}{\linewidth}{>{\raggedright\arraybackslash}X*{9}{c}>{\hspace{6pt}}c}
\toprule[0.8pt]
Method & GQA & MMB & MME & POPE & SQA & VQA$^{\text{v2}}$ & VQA$^{\text{Text}}$ & SEED$^{\text{I}}$ & VizWiz & RelAcc. \\
\midrule[0.4pt]
\multicolumn{11}{l}{\textcolor{black!60}{Uncompressed baseline (100\%)}} \\
Vanilla\textsubscript{\textcolor{black!60}{CVPR 2024}} & 61.9 & 64.7 & 1862 & 85.9 & 69.5 & 78.5 & 58.2 & 60.5 & 54.3 & 100\% \\
\specialrule{0.3pt}{3pt}{3pt}
\multicolumn{11}{l}{\textcolor{black!60}{Token retention: 33.3\%}} \\
FastV\textsubscript{\textcolor{black!60}{ECCV 2024}} & 52.7 & 61.2 & 1612 & 64.8 & 67.3 & 67.1 & 52.5 & 57.1 & 50.8 & 89.1\% \\
SparseVLM\textsubscript{\textcolor{black!60}{ICML 2025}} & 57.6 & 62.5 & 1721 & 83.6 & 69.1 & 75.6 & 56.1 & 55.8 & 50.5 & 95.2\% \\
DART\textsubscript{\textcolor{black!60}{EMNLP 2025}} & 60.0 & 63.6 & 1856 & 82.8 & 69.8 & 76.7 & 57.4 & 51.5 & 54.9 & 97.1\% \\
%ToMe\textsubscript{\textcolor{black!60}{ICLR 2023}} & 54.3 & 60.5 & 1563 & 72.4 & 65.2 & 68.0 & 52.1 & --- & --- & 88.5\% \\
%LLaVA-PruMerge\textsubscript{\textcolor{black!60}{ICCV 2025}} & 54.3 & 59.6 & 1632 & 71.3 & 67.9 & 70.6 & 54.3 & --- & 50.1 & 90.5\% \\
DivPrune\textsubscript{\textcolor{black!60}{CVPR 2025}} & 60.0 & 62.3 & 1752 & 87.0 & 68.7 & 75.5 & 56.4 & 58.6 & 55.6 & 97.8\% \\
VisionZip\textsubscript{\textcolor{black!60}{CVPR 2025}} & 59.3 & 63.0 & 1783 & 85.3 & 68.9 & 76.8 & 57.3 & 58.5 & 54.1 & 97.8\% \\
HiPrune\textsubscript{\textcolor{black!60}{ACL 2026}} & 59.2 & 62.8 & 1814 & 86.1 & 68.9 & 76.7 & 57.6 & --- & 54.5 & 98.3\% \\
EvoCut\textsubscript{\textcolor{black!60}{arXiv 2026.06}} & 60.2 & 64.2 & 1794 & 86.5 & 68.5 & 76.7 & 57.6 & 57.3 & --- & 97.9\% \\
PruneSID\textsubscript{\textcolor{black!60}{ICLR 2026}} & 60.1 & 63.7 & 1791 & 86.9 & 68.5 & 76.8 & 56.7 & 59.0 & 55.4 & \textbf{98.5\%} \\
\rowcolor{tableblue!60}
Ours & 60.1 & 63.6 & 1806 & 86.9 & 68.7 & 76.1 & 57.4 & 58.8 & 55.0 & \textbf{98.5\%} \\
\specialrule{0.3pt}{3pt}{3pt}
\multicolumn{11}{l}{\textcolor{black!60}{Token retention: 22.2\%}} \\
FastV\textsubscript{\textcolor{black!60}{ECCV 2024}} & 49.6 & 56.1 & 1490 & 59.6 & 60.2 & 61.8 & 50.6 & 55.9 & 51.3 & 83.9\% \\
SparseVLM\textsubscript{\textcolor{black!60}{ICML 2025}} & 56.0 & 60.0 & 1696 & 80.5 & 67.1 & 73.8 & 54.9 & 53.4 & 51.4 & 92.9\% \\
DART\textsubscript{\textcolor{black!60}{EMNLP 2025}} & 58.7 & 63.2 & 1840 & 80.1 & 69.1 & 75.9 & 56.4 & 50.5 & 55.3 & 95.9\% \\
%ToMe\textsubscript{\textcolor{black!60}{ICLR 2023}} & 52.4 & 53.3 & 1343 & 62.8 & 59.6 & 63.0 & 49.1 & --- & --- & 80.4\% \\
%LLaVA-PruMerge\textsubscript{\textcolor{black!60}{2024.05}} & 53.3 & 58.1 & 1554 & 67.2 & 67.1 & 68.8 & 54.3 & --- & 50.3 & 88.5\% \\
DivPrune\textsubscript{\textcolor{black!60}{CVPR 2025}} & 59.2 & 62.3 & 1752 & 86.9 & 69.0 & 74.7 & 56.0 & 57.1 & 55.6 & 97.2\% \\
VisionZip\textsubscript{\textcolor{black!60}{CVPR 2025}} & 57.6 & 62.0 & 1762 & 83.2 & 68.9 & 75.6 & 56.8 & 57.1 & 54.5 & 96.5\% \\
HiPrune\textsubscript{\textcolor{black!60}{ACL 2026}} & 57.3 & 62.2 & 1782 & 82.8 & 68.3 & 74.9 & 56.6 & --- & 54.3 & 96.5\% \\
EvoCut\textsubscript{\textcolor{black!60}{arXiv 2026.06}} & 59.2 & 62.3 & 1790 & 85.7 & 68.7 & 76.2 & 57.1 & 55.2 & --- & 96.6\% \\
PruneSID\textsubscript{\textcolor{black!60}{ICLR 2026}} & 58.8 & 62.1 & 1749 & 86.5 & 68.3 & 75.3 & 54.7 & 57.8 & 55.8 & 96.9\% \\
\rowcolor{tableblue!60}
Ours & 58.9 & 62.2 & 1778 & 86.7 & 68.7 & 75.1 & 56.8 & 57.7 & 56.0 & \textbf{97.6\%} \\
\specialrule{0.3pt}{3pt}{3pt}
\multicolumn{11}{l}{\textcolor{black!60}{Token retention: 11.1\%}} \\
FastV\textsubscript{\textcolor{black!60}{ECCV 2024}} & 46.1 & 48.0 & 1256 & 48.0 & 51.1 & 55.0 & 47.8 & 51.9 & 50.8 & 75.2\% \\
SparseVLM\textsubscript{\textcolor{black!60}{ICML 2025}} & 52.7 & 56.2 & 1505 & 75.1 & 62.2 & 68.2 & 51.8 & 51.1 & 53.1 & 87.5\% \\
DART\textsubscript{\textcolor{black!60}{EMNLP 2025}} & 55.9 & 60.6 & 1765 & 73.9 & 69.8 & 72.4 & 54.4 & 47.2 & 55.3 & 92.3\% \\
%ToMe\textsubscript{\textcolor{black!60}{ICLR 2023}} & 48.6 & 43.7 & 1138 & 52.5 & 50.0 & 57.1 & 45.3 & --- & --- & 70.1\% \\
%LLaVA-PruMerge\textsubscript{\textcolor{black!60}{2024.05}} & 51.9 & 55.3 & 1549 & 65.3 & 68.1 & 67.4 & 54.0 & --- & 50.1 & 87.2\% \\
DivPrune\textsubscript{\textcolor{black!60}{CVPR 2025}} & 57.6 & 59.3 & 1638 & 85.6 & 68.3 & 72.9 & 55.5 & 55.4 & 57.5 & 95.1\% \\
VisionZip\textsubscript{\textcolor{black!60}{CVPR 2025}} & 55.1 & 60.1 & 1690 & 77.0 & 69.0 & 72.4 & 55.5 & 54.5 & 54.8 & 93.4\% \\
HiPrune\textsubscript{\textcolor{black!60}{ACL 2026}} & 53.6 & 59.5 & 1646 & 73.0 & 68.9 & 69.2 & 54.9 & --- & 54.4 & 91.7\% \\
EvoCut\textsubscript{\textcolor{black!60}{arXiv 2026.06}} & 56.6 & 61.8 & 1692 & 83.9 & 68.8 & 73.4 & 55.7 & 53.2 & --- & 94.0\% \\
PruneSID\textsubscript{\textcolor{black!60}{ICLR 2026}} & 57.1 & 58.8 & 1733 & 83.8 & 67.8 & 73.7 & 54.2 & 56.1 & 56.9 & 95.1\% \\
\rowcolor{tableblue!60}
Ours & 57.7 & 59.5 & 1708 & 85.6 & 68.4 & 72.9 & 54.9 & 56.1 & 56.3 & \textbf{95.4\%} \\
\bottomrule[0.8pt]
\end{tabularx}
\endgroup
\end{table}

\subsection{Main results}
\label{sec:exp-image}
We follow the token retention settings of prior work~\citep{yang2025visionzip,fang2026prune}.
Across the four models, \ours{} achieves the highest or tied-highest RelAcc.\ among the compared methods at every reported budget.

\noindent\textbf{Results on LLaVA-1.5.}
We first evaluate \ours{} on LLaVA-1.5-7B~\citep{liu2024improved} across nine image understanding benchmarks, retaining 192, 128, and 64 of the original 576 visual tokens.
As shown in Table~\ref{tab:llava15-main}, \ours{} achieves RelAcc.\ of 98.5\%, 97.6\%, and 95.4\%, respectively.
When the budget is reduced to 64 tokens, \ours{} preserves 95.4\% of the baseline performance, compared with 95.1\% for both PruneSID and DivPrune and 93.4\% for VisionZip.
These results show that representation dynamics provide effective selection cues even when 88.9\% of the visual tokens are removed.

\noindent\textbf{Results on LLaVA-NeXT.}
To evaluate pruning with longer visual sequences, we apply \ours{} to LLaVA-NeXT-7B~\citep{liu2024llavanext}, which processes multiple image crops and produces up to 2,880 visual tokens.
Table~\ref{tab:llava-next} reports results at retention ratios of 22.2\%, 11.1\%, and 5.6\%.
Our method achieves RelAcc.\ of 97.5\%, 95.1\%, and 91.9\%, respectively, exceeding both VisionZip and PruneSID at all three budgets.
The advantage is particularly pronounced on POPE~\citep{li2023evaluating} under the smallest budget: with only 160 tokens, \ours{} obtains an F1 score of 87.1, compared with 76.9 for PruneSID and 74.8 for VisionZip.
This result highlights the effectiveness of the selected tokens in preserving evidence for object presence under severe compression.

\noindent\textbf{Results on Mini-Gemini.}
We next evaluate \ours{} on Mini-Gemini-7B~\citep{li2025mini} to examine its applicability to a dual-encoder architecture.
As shown in Table~\ref{tab:mini-gemini}, \ours{} achieves RelAcc.\ of 98.8\%, 98.0\%, and 95.2\% at 33.3\%, 22.2\%, and 11.1\% retention, respectively.
The gains over PruneSID increase from 0.3 to 0.9 and 2.1 percentage points as the token budget decreases, while \ours{} improve POPE F1 over PruneSID from 76.0 to 82.3 at 11.1\% retention.
The increasing advantage at smaller budgets shows that the proposed selection strategy remains effective when fewer tokens are available to represent the combined visual information.

\begin{table}[t]
\centering
\begin{minipage}[t]{0.5\linewidth}
\centering
\captionsetup{font=small,skip=4pt}
\caption{Performance on LLaVA-NeXT-7B.}
\label{tab:llava-next}
\setlength{\tabcolsep}{2.5pt}
\renewcommand{\arraystretch}{1.0}
\resizebox{\linewidth}{!}{%
\begin{tabular}{lcccccccc}
\toprule[0.8pt]
Method & GQA & MMB & MME & POPE & SQA & VQA$^{\text{v2}}$ & SEED$^{\text{I}}$ & RelAcc. \\
\midrule[0.4pt]
\multicolumn{9}{l}{\textcolor{black!60}{Uncompressed baseline (100\%)}} \\
Vanilla & 64.2 & 67.9 & 1842 & 86.4 & 70.2 & 80.1 & 70.2 & 100\% \\
\specialrule{0.3pt}{3pt}{3pt}
\multicolumn{9}{l}{\textcolor{black!60}{Token retention: 22.2\%}} \\
VisionZip & 61.3 & 66.3 & 1787 & 86.3 & 68.1 & 79.1 & 66.7 & 97.3\% \\
PruneSID & 61.6 & 64.2 & 1795 & 86.3 & 68.3 & 78.5 & 67.3 & 97.0\% \\
\rowcolor{tableblue!60}
Ours & 63.3 & 64.2 & 1805 & 87.6 & 68.0 & 77.6 & 67.5 & \textbf{97.5\%} \\
\specialrule{0.3pt}{3pt}{3pt}
\multicolumn{9}{l}{\textcolor{black!60}{Token retention: 11.1\%}} \\
VisionZip & 59.3 & 63.1 & 1702 & 82.1 & 67.3 & 76.2 & 63.4 & 93.4\% \\
PruneSID & 60.5 & 63.0 & 1754 & 83.1 & 67.3 & 76.6 & 65.0 & 94.6\% \\
\rowcolor{tableblue!60}
Ours & 61.0 & 63.1 & 1747 & 87.6 & 67.5 & 74.8 & 64.4 & \textbf{95.1\%} \\
\specialrule{0.3pt}{3pt}{3pt}
\multicolumn{9}{l}{\textcolor{black!60}{Token retention: 5.6\%}} \\
VisionZip & 55.5 & 60.1 & 1630 & 74.8 & 68.3 & 71.4 & 58.3 & 88.5\% \\
PruneSID & 58.9 & 60.8 & 1704 & 76.9 & 67.1 & 73.8 & 62.5 & 91.4\% \\
\rowcolor{tableblue!60}
Ours & 58.0 & 60.9 & 1627 & 87.1 & 68.5 & 71.5 & 60.9 & \textbf{91.9\%} \\
\bottomrule[0.8pt]
\end{tabular}%
}
\end{minipage}\hfill
\begin{minipage}[t]{0.5\linewidth}
\centering
\captionsetup{font=small,skip=4pt}
\caption{Performance on Mini-Gemini-7B.}
\label{tab:mini-gemini}
\setlength{\tabcolsep}{2.5pt}
\renewcommand{\arraystretch}{1.0}
\resizebox{\linewidth}{!}{%
\begin{tabular}{lcccccccc}
\toprule[0.8pt]
Method & GQA & MMB & MME & POPE & SQA & VQA$^{\text{v2}}$ & SEED$^{\text{I}}$ & RelAcc. \\
\midrule[0.4pt]
\multicolumn{9}{l}{\textcolor{black!60}{Uncompressed baseline (100\%)}} \\
Vanilla & 62.4 & 69.3 & 1841 & 85.8 & 70.7 & 80.4 & 69.7 & 100\% \\
\specialrule{0.3pt}{3pt}{3pt}
\multicolumn{9}{l}{\textcolor{black!60}{Token retention: 33.3\%}} \\
VisionZip & 60.3 & 68.9 & 1846 & 82.3 & 70.1 & 79.1 & 67.5 & 98.1\% \\
PruneSID & 61.2 & 67.2 & 1842 & 84.4 & 71.1 & 79.1 & 67.8 & 98.5\% \\
\rowcolor{tableblue!60}
Ours & 61.2 & 66.9 & 1884 & 85.4 & 71.5 & 78.0 & 67.4 & \textbf{98.8\%} \\
\specialrule{0.3pt}{3pt}{3pt}
\multicolumn{9}{l}{\textcolor{black!60}{Token retention: 22.2\%}} \\
VisionZip & 58.7 & 68.1 & 1841 & 78.5 & 70.0 & 77.5 & 65.6 & 96.2\% \\
PruneSID & 60.1 & 66.6 & 1821 & 82.4 & 70.7 & 77.8 & 66.5 & 97.1\% \\
\rowcolor{tableblue!60}
Ours & 60.5 & 67.4 & 1866 & 84.4 & 71.4 & 77.0 & 66.5 & \textbf{98.0\%} \\
\specialrule{0.3pt}{3pt}{3pt}
\multicolumn{9}{l}{\textcolor{black!60}{Token retention: 11.1\%}} \\
VisionZip & 55.8 & 65.9 & 1737 & 69.6 & 70.7 & 73.9 & 61.7 & 91.5\% \\
PruneSID & 58.3 & 63.1 & 1735 & 76.0 & 70.6 & 75.2 & 63.6 & 93.1\% \\
\rowcolor{tableblue!60}
Ours & 58.5 & 64.9 & 1782 & 82.3 & 71.6 & 74.7 & 64.0 & \textbf{95.2\%} \\
\bottomrule[0.8pt]
\end{tabular}%
}
\end{minipage}
\end{table}

\enlargethispage{11pt}
\vspace{-4pt}
\noindent\begin{minipage}[t]{0.49\linewidth}
\vspace{0pt}
\noindent\textbf{Results on Qwen2.5-VL.}
To assess generalization beyond the CLIP-based models, we evaluate \ours{} on Qwen2.5-VL-7B~\citep{bai2025qwen25vltechnicalreport}, which uses native dynamic resolution and spatial token merging.
As reported in Table~\ref{tab:qwen25vl}, \ours{} achieves RelAcc.\ of 97.2\%, 96.1\%, and 92.4\% at the three retention ratios, outperforming both compared methods throughout.
The gains over PruneSID increase from 0.4 to 0.9 and 1.5 percentage points as retention decreases.
Together with the results on the other models, these findings support the applicability of representation dynamics to visual token pruning across different encoder architectures and input-resolution schemes.

\end{minipage}%
\hfill
\begin{minipage}[t]{0.49\linewidth}
\vspace{-1pt}
\centering
\captionsetup{font=small,skip=4pt}
\captionof{table}{Qwen2.5-VL-7B (HRB$^{\text{8K}}$: HR-Bench 8K).}
\label{tab:qwen25vl}
\setlength{\tabcolsep}{3pt}
\renewcommand{\arraystretch}{1}
\resizebox{\linewidth}{!}{%
\begin{tabular}{lcccccccc}
\toprule[0.8pt]
Method & GQA & MMB & MME & POPE & SQA & VQA$^{\text{v2}}$ & HRB$^{\text{8K}}$ & RelAcc. \\
\midrule[0.4pt]
\multicolumn{9}{l}{\textcolor{black!60}{Uncompressed baseline (100\%)}} \\
Vanilla & 60.9 & 83.9 & 2310 & 86.3 & 88.9 & 82.9 & 68.1 & 100\% \\
\specialrule{0.3pt}{3pt}{3pt}
\multicolumn{9}{l}{\textcolor{black!60}{Token retention: 33.3\%}} \\
VisionZip & 56.6 & 78.9 & 2317 & 85.8 & 80.5 & 80.7 & 61.8 & 95.0\% \\
PruneSID & 59.8 & 80.9 & 2218 & 85.9 & 87.6 & 80.4 & 62.4 & 96.8\% \\
\rowcolor{tableblue!60}
Ours & 58.5 & 82.6 & 2332 & 84.7 & 87.8 & 80.5 & 61.9 & \textbf{97.2\%} \\
\specialrule{0.3pt}{3pt}{3pt}
\multicolumn{9}{l}{\textcolor{black!60}{Token retention: 22.2\%}} \\
VisionZip & 54.6 & 76.8 & 2224 & 83.4 & 80.4 & 78.5 & 61.3 & 92.8\% \\
PruneSID & 59.0 & 78.0 & 2169 & 85.6 & 86.9 & 78.7 & 61.8 & 95.2\% \\
\rowcolor{tableblue!60}
Ours & 57.7 & 81.2 & 2297 & 84.0 & 87.3 & 78.9 & 61.9 & \textbf{96.1\%} \\
\specialrule{0.3pt}{3pt}{3pt}
\multicolumn{9}{l}{\textcolor{black!60}{Token retention: 11.1\%}} \\
VisionZip & 53.2 & 75.8 & 2025 & 78.9 & 80.1 & 73.8 & 58.6 & 88.9\% \\
PruneSID & 55.8 & 73.9 & 2076 & 80.2 & 86.5 & 74.6 & 58.9 & 90.9\% \\
\rowcolor{tableblue!60}
Ours & 55.4 & 78.4 & 2178 & 81.3 & 86.3 & 75.0 & 59.0 & \textbf{92.4\%} \\
\bottomrule[0.8pt]
\end{tabular}%
}

\end{minipage}

\vspace{-4pt}
%\begin{comment}
\begin{wrapfigure}[10]{r}{0.49\linewidth}
\vspace{-1pt}
\centering
\captionsetup{font=small,skip=4pt}
\captionof{table}{Inference latency and POPE F1 score on LLaVA-NeXT-7B. Latencies are averaged per sample using a single NVIDIA A800-80\,GB GPU.}
\label{tab:efficiency}
\begingroup
\fontsize{9}{10}\selectfont
\setlength{\tabcolsep}{9pt}
\renewcommand{\arraystretch}{1}
\resizebox{\linewidth}{!}{%
\begin{tabular}{lcccc}
\toprule[0.8pt]
\raisebox{0.5\baselineskip}{Method} & \raisebox{0.5\baselineskip}{Tokens} &
\begin{tabular}{@{}c@{}}Inference\\Time~$\downarrow$\end{tabular} &
\begin{tabular}{@{}c@{}}Prefilling\\Time~$\downarrow$\end{tabular} &
\begin{tabular}{@{}c@{}}POPE\\(F1)~$\uparrow$\end{tabular} \\
\midrule[0.4pt]
LLaVA-NeXT & 2880 & 254\,ms & 218\,ms & 86.4 \\
\midrule[0.4pt]
SparseVLM & 160 & 199\,ms & 119\,ms & 69.3 \\
VisionZip & 160 & \textbf{84\,ms} & \textbf{27.8\,ms} & 74.8 \\
PruneSID & 160 & 89\,ms & \textbf{27.8\,ms} & 76.9 \\
\rowcolor{tableblue!60}
Ours & 160 & 96\,ms & \textbf{27.8\,ms} & \textbf{87.1} \\
\bottomrule[0.8pt]
\end{tabular}%
}
\endgroup
\end{wrapfigure}

\noindent\textbf{Inference Efficiency.}
Following prior work~\citep{fang2026prune,yang2025visionzip}, we measure per-sample latency on LLaVA-NeXT-7B using a single NVIDIA A800-80\,GB GPU.
As shown in Table~\ref{tab:efficiency}, retaining 160 tokens reduces prefilling time from 218\,ms to 27.8\,ms and total inference time from 254\,ms to 96\,ms, yielding speedups of 7.8$\times$ and 2.6$\times$, respectively.
POPE-F1 score reaches 87.1, compared with 86.4 without pruning.
Compared with PruneSID, \ours{} matches prefilling latency and gains 10.2 percentage points with 7\,ms additional total latency, balancing inference efficiency and object grounding.

\vspace{-4pt}

\subsection{Ablation study}
\label{sec:exp-ablation}
As shown in Table~\ref{tab:ablation-signals}, we conduct component ablations on Qwen2.5-VL-7B to examine the contributions of the grouping representation, visual saliency, sink filtering, and budget allocation.

\noindent\textbf{Representation and Saliency Signals.}
Clustering with update directions $d_i$ outperforms grouping with encoder-output features at all retention ratios.
Together with the analysis in Section~\ref{sec:update-direction-semantic-consistency}, 
this supports using representation changes to form semantically coherent groups.
For visual saliency, $u_i$ matches mean received attention in the final encoder layer at 33.3\% retention and improves RelAcc.\ by 0.4 and 0.6 points at the smaller budgets.
Compared with accumulated update magnitude, computing $u_i$ from endpoint displacement yields gains of 0.2--1.4 points on Qwen2.5-VL, supporting net representation change as the saliency cue.
The effect of the group count $K$ is evaluated in Appendix~\ref{app:ablation-k}.
These results support our design choices for the representation and saliency signals.

\begin{table}[b]
\centering
\captionsetup{font=small,skip=4pt}
\caption{Component ablations and staged token selection on Qwen2.5-VL-7B.
Ours is the full method; Output features uses encoder-output features for grouping.
Query (global) ranks by query relevance without grouping;
Query$\times$Visual (global) ranks by the product of query relevance and visual saliency, still without grouping.}
\label{tab:qwen-saliency-score}
\label{tab:qwen-update-accum}
\label{tab:ablation-signals}
\label{tab:ablation-cluster}
\label{tab:ablation-pipeline}
\setlength{\tabcolsep}{2.5pt}
\renewcommand{\arraystretch}{0.96}
\resizebox{\linewidth}{!}{%
\begin{tabular}{lrrrrrrrrrrrrrrr}
\toprule[0.8pt]
\multirow{2}{*}{Method} & \multicolumn{5}{c}{\textcolor{black!60}{Retain 33.3\%}}
& \multicolumn{5}{c}{\textcolor{black!60}{Retain 22.2\%}}
& \multicolumn{5}{c}{\textcolor{black!60}{Retain 11.1\%}} \\
\cmidrule(lr){2-6}\cmidrule(lr){7-11}\cmidrule(lr){12-16}
& GQA & MME & POPE & SQA & RelAcc.
& GQA & MME & POPE & SQA & RelAcc.
& GQA & MME & POPE & SQA & RelAcc. \\
\midrule[0.4pt]
\rowcolor{tableblue!60}
Ours & 58.5 & 2332 & 84.7 & 87.8 & \textbf{98.5\%}
& 57.7 & 2297 & 84.0 & 87.3 & \textbf{97.4\%}
& 55.4 & 2178 & 81.3 & 86.3 & \textbf{94.1\%} \\
\specialrule{0.3pt}{3pt}{3pt}
Output features & 58.1 & 2286 & 84.0 & 87.5 & 97.5\%
& 56.9 & 2254 & 82.8 & 87.1 & 96.2\%
& 54.8 & 2178 & 80.1 & 86.0 & 93.5\% \\
Attention & 57.9 & 2319 & 86.1 & 87.9 & \textbf{98.5\%}
& 56.9 & 2268 & 84.4 & 87.6 & 97.0\%
& 53.8 & 2176 & 81.3 & 86.5 & 93.5\% \\
Accumulated updates & 58.5 & 2317 & 84.4 & 87.9 & 98.3\%
& 56.5 & 2272 & 83.2 & 87.3 & 96.4\%
& 54.3 & 2129 & 80.1 & 85.9 & 92.7\% \\
Keep sink tokens & 58.4 & 2330 & 84.7 & 87.7 & 98.4\%
& 57.1 & 2264 & 83.1 & 87.2 & 96.5\%
& 54.8 & 2143 & 81.2 & 86.3 & 93.5\% \\
Size-based budget & 58.8 & 2281 & 84.4 & 87.8 & 98.0\%
& 57.9 & 2247 & 83.3 & 87.2 & 96.7\%
& 55.4 & 2111 & 80.2 & 85.9 & 93.0\% \\
\specialrule{0.3pt}{3pt}{3pt}
Random & 55.2 & 2275 & 82.3 & 87.3 & 95.7\%
& 56.9 & 2166 & 81.1 & 86.6 & 94.6\%
& 53.6 & 1995 & 77.4 & 85.2 & 90.0\% \\
Query (global) & 56.1 & 2247 & 80.9 & 87.3 & 95.3\%
& 54.6 & 2146 & 77.7 & 86.9 & 92.6\%
& 51.4 & 1935 & 71.5 & 85.1 & 86.7\% \\
Query$\times$Visual (global) & 58.3 & 2337 & 84.8 & 87.8 & \textbf{98.5\%}
& 56.6 & 2282 & 82.2 & 87.2 & 96.3\%
& 52.4 & 2152 & 81.1 & 86.1 & 92.5\% \\
\bottomrule[0.8pt]
\end{tabular}%
}
\end{table}

%$u_i^{\mathrm{acc}}=\sum_{\ell=\ell_s}^{\ell_e-1}\lVert h_i^{\ell+1}-h_i^\ell\rVert$
\noindent\textbf{Sink Filtering and Budget Allocation.} 
We examine the contribution of sink removal and budget allocation.
Retaining sink tokens reduces RelAcc.\ at all retention ratios, with a drop of 0.9 percentage points at 22.2\% retention.
This result supports sink filtering.
Instead of using group scores, allocating budgets in proportion to group size also lowers RelAcc.\ by 0.5--1.1 percentage points.
The largest reduction occurs at 11.1\% retention, highlighting the importance of allocating limited tokens according to group importance.

\noindent\textbf{Complementarity of Scoring Signals.}
We compare random selection, global ranking by query relevance, global ranking by query-weighted visual saliency, and the full method.
Query relevance alone performs below random selection at all three retention ratios, showing that similarity to the query embeddings is insufficient as a standalone selection criterion.
For global token ranking, combining query relevance with visual saliency improves RelAcc.\ by 3.2--5.8 percentage points over query relevance alone across the three retention ratios.
Group-wise pruning further improves over global ranking with the combined score by 1.1 and 1.6 points at 22.2\% and 11.1\% retention, respectively.
%This supports visual saliency as a complementary cue to query relevance for token selection, while grouping improves budget allocation across visual content.

\noindent\textbf{Saliency Window Endpoints.}\ %包含每个阶段的代表选择点
We compare saliency windows spanning five consecutive layers at different encoder depths.
The default windows achieve the highest RelAcc.\ among the configurations in Figure~\ref{fig:window-summary} at both retention ratios.
Using the final five encoder updates instead reduces RelAcc.\ by 3.7 and 5.8 percentage points on Qwen2.5-VL, and by 2.0 and 2.2 points on LLaVA-1.5, respectively.
This degradation is consistent with the increased background responses near the encoder output, supporting saliency estimation within the late foreground-enhanced stage.

\vspace{10pt}
\FloatBarrier
\begin{minipage}[t]{0.48\linewidth}
\centering
\includegraphics[width=\linewidth]{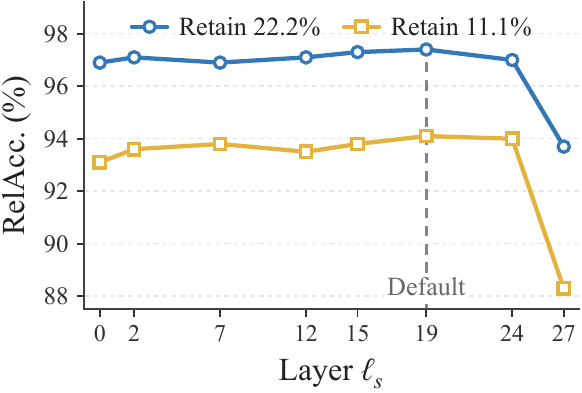}\\[-1pt]
{\small (a) Qwen2.5-VL}
\end{minipage}\hfill
\begin{minipage}[t]{0.48\linewidth}
\centering
\includegraphics[width=\linewidth]{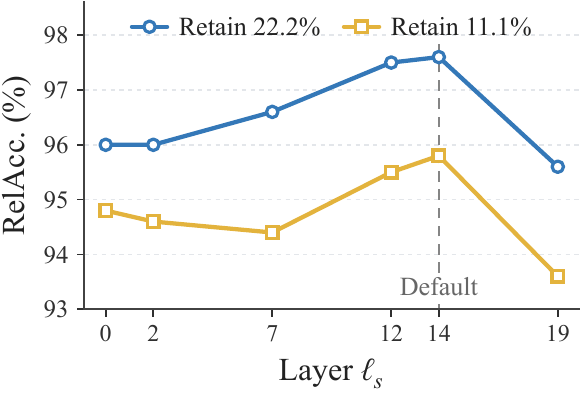}\\[-1pt]
{\small (b) LLaVA-1.5}
\end{minipage}

\vspace{2pt}
\captionsetup{font=small,skip=4pt}
\captionof{figure}{Effect of saliency window start on RelAcc.\ for (a) Qwen2.5-VL-7B and (b) LLaVA-1.5-7B at 22.2\% and 11.1\% token retention. Window starts $\ell_s$ are selected as representative points across the stages identified in the preceding analysis. Each window spans five consecutive encoder updates. Dashed lines mark the default starts.}
\label{fig:window-summary}
\subsection{Limitations}
\label{sec:limitations}
Because \ours{} requires intermediate vision encoder states, it cannot be applied through interfaces that expose only final predictions.
Moreover, generalizing our method to new backbones entails analyzing and identifying specific model-dependent components, which requires an in-depth understanding of the model architecture.

\section{Conclusion}
\label{sec:conclusion}

We present \ours{}, a training-free visual token pruning method guided by encoder representation dynamics. It uses update directions for grouping and combines update magnitudes with query relevance for saliency. At 11.1\% token retention, \ours{} achieves 95.4\% and 92.4\% RelAcc.\ on LLaVA-1.5-7B and Qwen2.5-VL-7B, respectively. On LLaVA-NeXT-7B, it achieves a 7.8$\times$ prefilling speedup at 5.6\% retention.

\subsection*{AI use statement}
In this work, we used generative AI tools (Cursor, Codex) to generate and debug experimental code and evaluation scripts for selected benchmarks, and to edit the manuscript to improve English readability and polish presentation (e.g., phrasing, clarity, and \LaTeX{} formatting).
We did not use AI tools for research design, theoretical or mathematical claims, data analysis, or interpretation of results.
The authors reviewed AI-assisted text, tested AI-assisted code, and obtained all reported results by running the evaluations.
The authors take responsibility for the final content.

\subsection*{Ethics statement}
The research presented in this paper focuses on visual token pruning for MLLMs to improve their computational efficiency, and does not involve human subjects, personally identifiable information, or the collection of new datasets.
The research process of this paper does not violate the ICLR Code of Ethics.
We do not train new models; our method is an inference-time pruning technique applied to existing publicly available MLLMs.
We do not foresee additional discrimination, bias, or fairness concerns beyond those already inherent in the underlying base models, nor do we expect our method to introduce new harmful generation capabilities.

\subsection*{Reproducibility statement}
The base models and datasets used in our experiments are all publicly available and cited, which supports reproducibility of the evaluation setting.
To preserve anonymity during double-blind review, we do not release source code at submission time; upon acceptance, we will publicly release the source code to support reproducibility.

\bibliography{iclr2027_conference}
\bibliographystyle{iclr2027_conference}

\clearpage
\appendix
\noindent\textbf{Appendix Overview.} For convenient reference, the appendix is organized as follows:
\begin{itemize}[leftmargin=1.3em,topsep=3pt,itemsep=1pt,parsep=0pt]
    \item \textbf{Appendix~\ref{app:extra}: Additional Algorithm Details}
    \item \textbf{Appendix~\ref{app:feature-dynamics-clustering}: Additional Analysis of Representation Dynamics}
    \begin{itemize}[label=--,leftmargin=1.5em,topsep=0pt,itemsep=0pt,parsep=0pt]
        \item Update magnitude analysis
        \item Update direction analysis
    \end{itemize}
    \item \textbf{Appendix~\ref{app:extra-exp}: Additional Experimental Results}
    \begin{itemize}[label=--,leftmargin=1.5em,topsep=0pt,itemsep=0pt,parsep=0pt]
        \item Implementation settings
        \item Additional evaluation on larger models
        \item Ablations on designed components
    \end{itemize}
    \item \textbf{Appendix~\ref{app:benchmarks}: Details of Evaluation Benchmarks}
    \item \textbf{Appendix~\ref{app:viz}: Qualitative Analysis of Retained Tokens}
    \begin{itemize}[label=--,leftmargin=1.5em,topsep=0pt,itemsep=0pt,parsep=0pt]
        \item Failure cases
        \item Comparison of retained tokens
    \end{itemize}
\end{itemize}

\section{Additional Algorithm Details}
\label{app:extra}
\label{app:algo-eff}
We detail the token selection algorithm, centroid initialization, and stopping criteria, show that the grouping objective is equivalent to minimizing squared chord distance, and relate endpoint displacement to the accumulated update magnitude evaluated in Section~\ref{sec:exp-ablation}.

\newcounter{algorithm}
\begin{table}[hb]
\centering
\refstepcounter{algorithm}
\label{alg:algorithm}
\small
\setlength{\tabcolsep}{2pt}
\renewcommand{\arraystretch}{1.02}
\begin{tabular}{@{}r@{\hspace{0.6em}}>{\raggedright\arraybackslash}p{0.925\textwidth}@{}}
\toprule
\multicolumn{2}{@{}l}{\textbf{Algorithm \thealgorithm}\quad\textbf{\ours{} Visual Token Selection}} \\
\midrule
\multicolumn{2}{@{}>{\raggedright\arraybackslash}p{0.98\textwidth}@{}}{\textbf{Require:} encoder states $\{h_i^\ell:1\leq i\leq N,\ 0\leq\ell \leq L\}$,
projected visual tokens $\{\bm{x}_i\}_{i=1}^{N}$, where $N$ is the number of visual tokens before pruning and $L$ is the number of encoder layers,
query embeddings $\{\bm{t}_j\}_{j\in\mathcal Q}$, where $\mathcal Q$ indexes the query tokens,
budget $B$, group count $K$,
$\ell_0$ and $\ell_d$ (start and end layers of the update direction $d_i$),
$\ell_\star$ (peak concentration layer; Appendix~\ref{app:sink-detection}),
$\ell_s$ and $\ell_e$ (start and end of the saliency window for the visual saliency score),
diagnostic coordinate $d_s$, threshold $\tau$, and integer initialization seed $s$;
$\operatorname{unit}(z)=z/\max\{\lVert z\rVert,10^{-12}\}$.
$\operatorname{TopIdx}_b$ returns the indices of the $b$ largest scores, breaking ties by original token order.} \\
1: & $\mathcal V\leftarrow\{i: |(h_i^{\ell_\star+1})_{d_s}|\leq\tau\}$; require $B\leq|\mathcal V|$, $1\leq K\leq|\mathcal V|$, and $\mathcal Q\neq\varnothing$ \\
2: & \textbf{for each} token $i\in\mathcal V$ \textbf{do} \\
3: & \quad$\Delta h_i\leftarrow\operatorname{unit}(h_i^{\ell_d})-
\operatorname{unit}(h_i^{\ell_0})$; $d_i\leftarrow\operatorname{unit}(\Delta h_i)$ \\
4: & \quad$u_i\leftarrow\lVert h_i^{\ell_e}-h_i^{\ell_s}\rVert$;
$\alpha_i\leftarrow\max_{j\in\mathcal Q}
\operatorname{unit}(\bm{x}_i)^\top\operatorname{unit}(\bm{t}_j)$ \\
5: & \textbf{end for} \\
6: & Set $s_i\leftarrow\alpha_i u_i$ for each $i\in\mathcal V$ \\
7: & Require $\max_{i\in\mathcal V}\lVert d_i\rVert>0$; initialize centroids $\{\mu_c\}_{c=1}^{K}$ by deterministic farthest-point sampling \\
8: & \textbf{for} $t=1,\ldots,10$ \textbf{do} \\
9: & \quad Assign each $i\in\mathcal V$ to a centroid maximizing $d_i^\top\mu_c$; set $m_i\leftarrow\max_c d_i^\top\mu_c$, form $\mathcal C_c$, and compute $\mathcal L^{(t)}$ \\
10: & \quad\textbf{if} $t>1$ and (labels are unchanged or the relative loss change is below $10^{-5}$) \textbf{then break} \\
11: & \quad Set $\mu_c\leftarrow\operatorname{unit}(\sum_{i\in\mathcal C_c}d_i)$ for each nonempty $\mathcal C_c$ \\
12: & \quad For empty groups, set $\mu_c$ to distinct $d_i$ in ascending $m_i$ order (ties: original token order) \\
13: & \textbf{end for}; reassign tokens to the final centroids to obtain $\{\mathcal C_c\}_{c=1}^{K}$ \\
14: & Set $g_c\leftarrow |\mathcal C_c|^{-1}\sum_{i\in\mathcal C_c}s_i$ if $|\mathcal C_c|>0$, and $g_c\leftarrow0$ otherwise \\
15: & $p_c\leftarrow\exp(g_c)/\sum_{r=1}^{K}\exp(g_r)$ for $c=1,\ldots,K$ \\
16: & $b_c\leftarrow\min\{|\mathcal C_c|,\lfloor Bp_c\rfloor\}$ for $c=1,\ldots,K$; $R\leftarrow B-\sum_{c=1}^{K}b_c$ \\
17: & Visit groups with $b_c<|\mathcal C_c|$ in descending order of $Bp_c-\lfloor Bp_c\rfloor$ (ties: lower $c$) \\
18: & \quad While $R>0$, add one token to each visited group: $b_c\leftarrow b_c+1$, $R\leftarrow R-1$ \\
19: & Visit groups in descending order of $p_c$ (ties: lower $c$); set $\Delta b_c\leftarrow\min\{R,|\mathcal C_c|-b_c\}$, $b_c\leftarrow b_c+\Delta b_c$, $R\leftarrow R-\Delta b_c$ \\
20: & $\mathcal S_c\leftarrow\operatorname{TopIdx}_{b_c}(\{s_i\}_{i\in\mathcal C_c})$ for each $c=1,\ldots,K$ \\
21: & $\mathcal S\leftarrow\operatorname{Sort}(\bigcup_{c=1}^{K}\mathcal S_c)$ by the original visual token indices \\
22: & \textbf{return} $\widetilde{\bm{X}}\leftarrow[\bm{x}_i]_{i\in\mathcal S}$ \\
\bottomrule
\end{tabular}
\end{table}
\paragraph{Visual token selection.}
Algorithm~\ref{alg:algorithm} summarizes the visual token selection procedure of \ours{}.
Update directions and visual saliency scores are computed from vision encoder states, while query relevance is measured between projected visual tokens and query embeddings in the LLM embedding space.
Sink tokens are excluded from grouping, group scoring, and token selection.

\paragraph{Deterministic centroid initialization and stopping.}
We initialize the centroids by deterministic farthest-point sampling using inner products of unit update directions.
Let $M=|\mathcal V|$, and let $v_0,\ldots,v_{M-1}$ list the candidate token indices in their original order.
The integer seed $s$ determines the first index $i_1=v_{s\bmod M}$ and centroid $\mu_1=d_{i_1}$.
Each subsequent centroid is selected as
\begin{equation}
    i_c=\arg\max_{i\in\mathcal V\setminus\{i_1,\ldots,i_{c-1}\}}
    \left(1-\max_{1\leq r<c}d_i^\top\mu_r\right),
    \qquad \mu_c=d_{i_c},\quad c=2,\ldots,K.
    \label{eq:farthest-sampling}
\end{equation}
This rule selects an unchosen token with the largest cosine distance to its nearest existing centroid, breaking ties by original token order.
At iteration \(t\), let \(a_i^{(t)}\) denote the group label of token \(i\), and let \(\mu_c^{(t)}\) denote centroid \(c\).
The grouping loss is the mean cosine distance to the assigned centroids,
\begin{equation}
    \mathcal L^{(t)}=\frac{1}{M}\sum_{i\in\mathcal V}
    \left(1-d_i^\top\mu_{a_i^{(t)}}^{(t)}\right).
    \label{eq:clustering-loss}
\end{equation}
Minimizing this loss is equivalent to maximizing the cosine alignment objective in Equation~\ref{eq:group-wise-clustering-draft}.
The stopping criterion based on relative loss change is
\begin{equation}
    \frac{\left|\mathcal L^{(t-1)}-\mathcal L^{(t)}\right|}
    {\max\!\left(\left|\mathcal L^{(t-1)}\right|,10^{-12}\right)}
    <10^{-5}.
    \label{eq:clustering-stop}
\end{equation}
Each grouping run stops when the assignment labels remain unchanged, the relative loss change satisfies Equation~\ref{eq:clustering-stop}, or $10$ iterations have been completed.
The label and loss comparisons are evaluated from the second iteration onward.

\paragraph{Equivalent grouping objective.}
In the nondegenerate case where every $d_i$ and $\mu_c$ has unit norm, any partition of the candidate tokens satisfies
\begin{equation}
\begin{split}
    \sum_{c=1}^{K}\sum_{i\in\mathcal C_c}
    \lVert d_i-\mu_c\rVert^2
    &=\sum_{c=1}^{K}\sum_{i\in\mathcal C_c}
    \left(2-2d_i^\top\mu_c\right) \\
    &=2|\mathcal V|-2\sum_{c=1}^{K}\sum_{i\in\mathcal C_c}
    d_i^\top\mu_c.
    \label{eq:sphericalequiv}
\end{split}
\end{equation}
Thus, maximizing the objective in Equation~\ref{eq:group-wise-clustering-draft} is equivalent to minimizing
the total squared chord distance between unit update directions and unit centroids.
The grouping objective therefore depends only on the unit update directions.

\paragraph{Endpoint displacement and accumulated update magnitude.}
Let $\delta_i^\ell=h_i^{\ell+1}-h_i^\ell$ be the update through layer $\ell$.
The visual saliency score in Section~\ref{sec:foreground-window-saliency-estimation} satisfies
\begin{equation}
    u_i
    =\lVert h_i^{\ell_e}-h_i^{\ell_s}\rVert
    =\left\lVert\sum_{\ell=\ell_s}^{\ell_e-1}\delta_i^\ell\right\rVert
    \leq
    \sum_{\ell=\ell_s}^{\ell_e-1}\lVert\delta_i^\ell\rVert
    =\sum_{\ell=\ell_s}^{\ell_e-1}v_i^\ell.
    \label{eq:pathbound}
\end{equation}
The second equality follows from telescoping, and the inequality follows from the triangle inequality.
Equality holds when all nonzero updates point in the same direction.
Endpoint displacement measures the representation change between the window endpoints, allowing opposing updates to cancel.
The accumulated update magnitude measures the total length of the representation path within the window.
The ablations in Tables~\ref{tab:qwen-update-accum} and~\ref{tab:llava-update-accum} compare these two aggregation rules in downstream compression.

\clearpage
\section{Additional Analysis of Representation Dynamics}
\label{app:feature-dynamics-clustering}

\subsection{Update Magnitude Analysis}
\label{app:feature-dynamics}
We report layer-wise update statistics, evaluate foreground recall using visual saliency scores $u_i$ computed from endpoint displacement, and visualize spatial update patterns for both vision encoders.

\begin{figure}[!b]
\centering
\begin{minipage}[t]{0.48\linewidth}
\centering
\includegraphics[width=\linewidth]{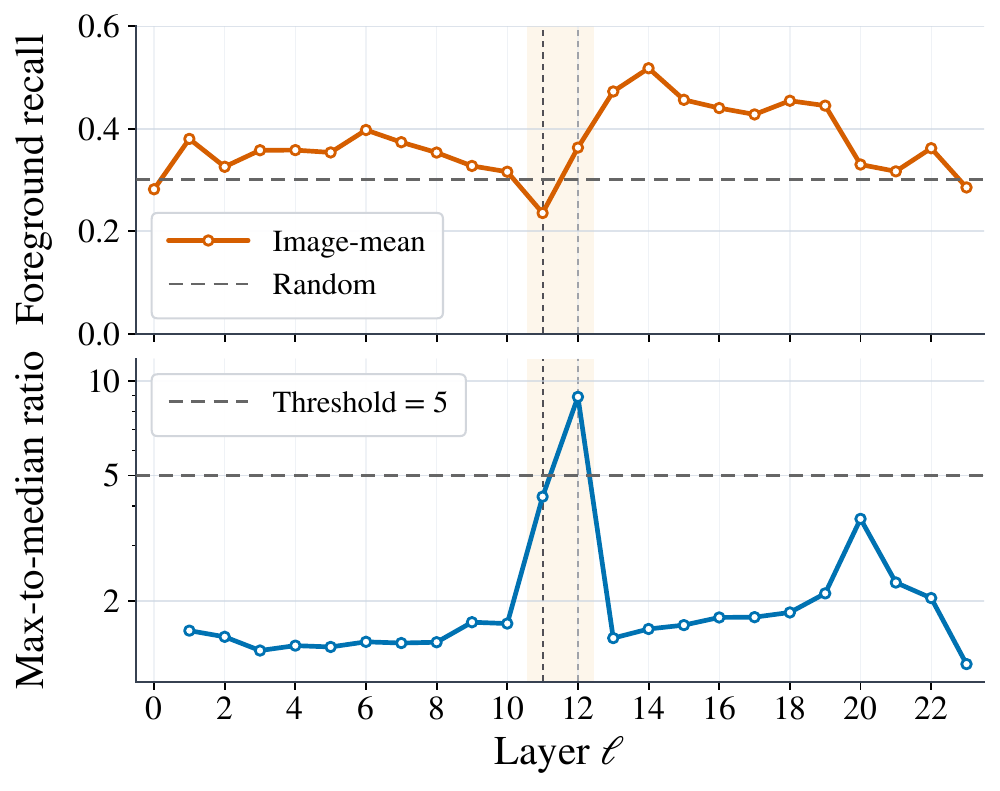}\\[2pt]
{\small (a) LLaVA-NeXT}
\end{minipage}\hfill
\begin{minipage}[t]{0.48\linewidth}
\centering
\includegraphics[width=\linewidth]{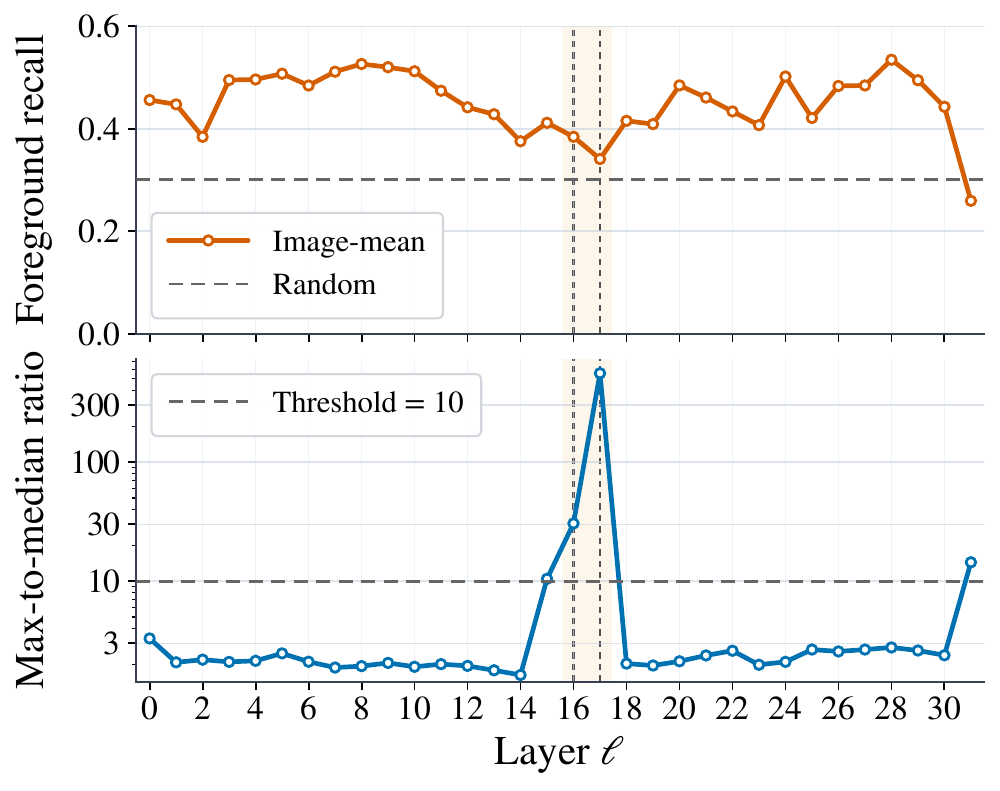}\\[2pt]
{\small (b) Qwen2.5-VL}
\end{minipage}
\caption{Cross-image sink signature on COCO 2017 validation images on LLaVA-NeXT and Qwen2.5-VL.
Top: image-mean layer-wise foreground recall when selecting the top $30\%$ of tokens by $v_i^\ell$.
Bottom: mean concentration ratio $\max_i v_i^\ell/\operatorname{median}_i v_i^\ell$; the dashed line marks the threshold ($5$ on LLaVA-NeXT and $10$ on Qwen2.5-VL).
A ratio spike at a fixed intermediate depth indicates isolated tokens with large update magnitudes, consistent with the sink-dominated stage.}
\label{fig:app-sink-ratio}
\end{figure}

\paragraph{Cross-image consistency of the sink-dominated stage.}
\label{app:sink-consistency}
To assess whether the sink-dominated stage generalizes beyond the visualized examples, we compute two layer-wise statistics on the COCO 2017 validation set ($4{,}952$ valid images): image-mean foreground recall when selecting the top $30\%$ of tokens by $v_i^\ell$, and the concentration ratio $\max_i v_i^\ell/\operatorname{median}_i v_i^\ell$.
Figure~\ref{fig:sink-ratio-llava} reports these quantities for LLaVA-1.5; Figure~\ref{fig:app-sink-ratio} reports the same diagnostics on LLaVA-NeXT and Qwen2.5-VL, with thresholds $5$ and $10$, respectively.
A ratio above the dashed threshold indicates that the largest updates are isolated to a few tokens rather than distributed across the image, matching the spatial pattern of the sink-dominated stage.

On CLIP-ViT-L~\citep{radford2021learning}, the mean ratio exceeds the threshold only at Layers 11 and 12 and remains near $2$ at other depths.
Foreground recall at these layers is close to that of random selection, consistent with isolated sink updates rather than contiguous foreground change.
On Qwen2.5-VL, the same isolated-token spike is confined to Layers 15--17.
A much smaller rise appears near the encoder output and does not reproduce the intermediate concentration pattern.
The sharpness of these image-mean peaks implies a shared sink-dominated depth within each encoder: if the peak layer varied widely across images, the average ratio would be smeared.
We use the peak concentration layer $\ell_\star$, identified in Appendix~\ref{app:sink-detection}, as a fixed reference for window placement within each encoder.
The transition occurs around these reference layers, with some variation across images.
In both encoders, the subsequent rise in foreground recall motivates us to explore this late foreground-enhanced stage for constructing visual saliency scores.

\paragraph{Foreground recall of fixed-width windows.}
\label{app:window-foreground-recall}
We fix the window width to five updates and scan its start on the same COCO 2017 validation images for both encoders, using COCO instance annotations~\citep{lin2014microsoft}.
For each start $\ell_s$, we rank eligible patches by the visual saliency score $u_i=\lVert h_i^{\ell_e}-h_i^{\ell_s}\rVert$, with $\ell_e=\ell_s+5$, and select the top $30\%$.
For image $q$, let $\mathcal T_{q,\ell_s}$ be the selected patch set, $P_{q,i}$ the pixel region of patch $i$, and $M_{q,j}$ the mask of the $j$-th evaluated instance.
The recall for this instance is
\begin{equation}
    R_{q,j}(\ell_s)=\frac{\sum_{i\in\mathcal T_{q,\ell_s}}|P_{q,i}\cap M_{q,j}|}{|M_{q,j}|}.
    \label{eq:foreground-area-recall}
\end{equation}
We first average instance recalls within each image, then average the resulting image scores over the $4{,}952$ images.
Thus, each image contributes equally to the reported metric, and instances within an image receive equal weight.

\begin{figure}[tbp]
\centering
\includegraphics[width=\linewidth]{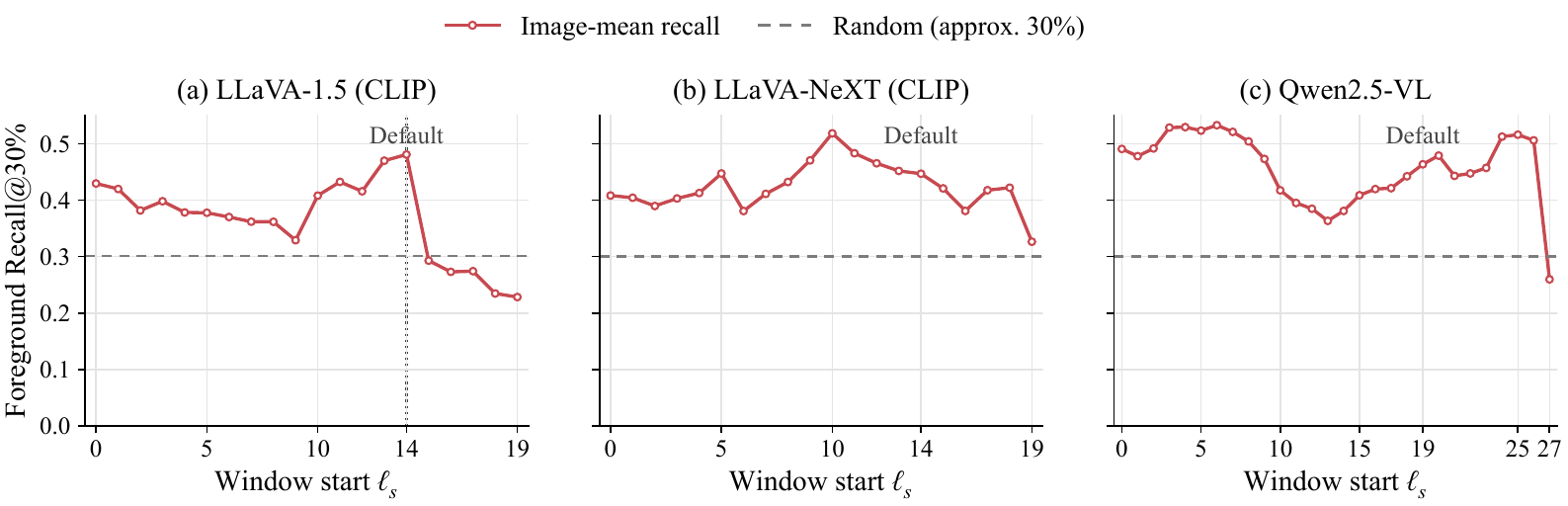}
\caption{Image-mean Foreground Recall@30\% for windows of width $w=5$ on COCO 2017 validation images.
At each input-state start $\ell_s$, patches are ranked by $\lVert h_i^{\ell_s+5}-h_i^{\ell_s}\rVert$.
Horizontal dashed lines show the random selection expectation (approximately $30\%$); vertical dotted lines mark the default starts used in the main experiments.}
\label{fig:window-foreground-recall}
\end{figure}

CLIP images are padded to a square and resized to $336\times336$, with [CLS] and padding tokens excluded from selection.
For Qwen2.5-VL, we apply standard dynamic resizing and evaluate recall on $14\times14$ image patches before spatial merging.
We undo the token permutation used for window attention to map each patch score back to its image location for comparison with the instance masks.
Sink filtering is disabled in this diagnostic.
Masks are resized with nearest-neighbor interpolation.
Figure~\ref{fig:window-foreground-recall} reports the mean foreground recall across images when the top $30\%$ of patches are selected by the visual saliency score $u_i$.

All plotted starts follow the input-state convention of Section~\ref{sec:representation-dynamics}, so the default windows appear at $\ell_s=14$ and $\ell_s=19$.
On LLaVA-1.5, the default $14\to19$ window attains the highest image-mean recall, $48.1\%$.
On Qwen2.5-VL, the default $19\to24$ window reaches $46.4\%$, well above the random baseline of about $30\%$, while the shallow $2\to7$ window reaches $49.2\%$.
The default Qwen2.5-VL window yields higher downstream RelAcc.\ than $2\to7$ in Table~\ref{tab:qwen-window}.
Foreground recall alone therefore does not determine which window works best in the full pruning pipeline, which also uses semantic grouping and query relevance.
The scan evaluates window-level foreground coverage; the separate sink analysis above provides the reference for its placement.

\paragraph{Layer-wise update visualizations.}
Figures~\ref{fig:app-tennis}--\ref{fig:app-tennis-qwen} present additional layer-wise visualizations of token update magnitudes $v_i^\ell$.
Figure~\ref{fig:app-tennis} shows the tennis scene for LLaVA-1.5 and Mini-Gemini, since both models use the same CLIP-ViT-L encoder~\citep{radford2021learning}.
Figures~\ref{fig:app-kite-next} and~\ref{fig:app-kite-qwen} show the kite scene for LLaVA-NeXT and Qwen2.5-VL.
Figures~\ref{fig:app-tennis-next} and~\ref{fig:app-tennis-qwen} show the tennis scene for LLaVA-NeXT and Qwen2.5-VL.
Each panel maps update magnitudes to their corresponding spatial positions in the input image; a label $\ell\to\ell+1$ denotes the update through Layer $\ell$.
These examples illustrate how the spatial distribution of updates varies across encoder depth for different images and encoders.

\begingroup
\captionsetup{skip=4pt}
\begin{figure}[!b]
\centering
\includegraphics[width=0.94\linewidth]{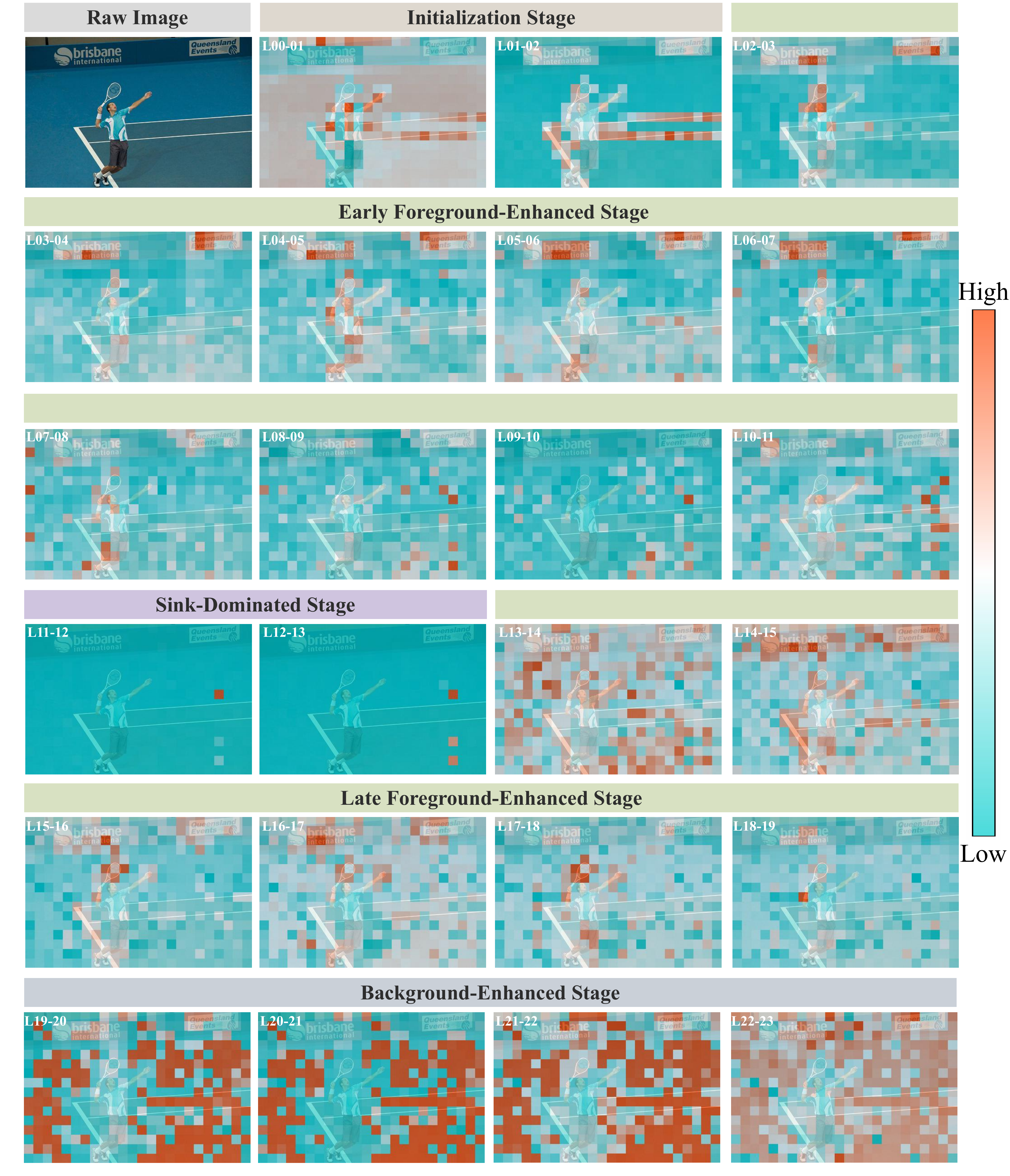}
\caption{Spatial maps of update magnitudes $v_i^{\ell}$ across LLaVA-1.5 and Mini-Gemini vision encoder layers for the tennis scene.}
\label{fig:app-tennis}
\end{figure}

\begin{figure}[tbp]
\centering
\includegraphics[width=0.94\linewidth]{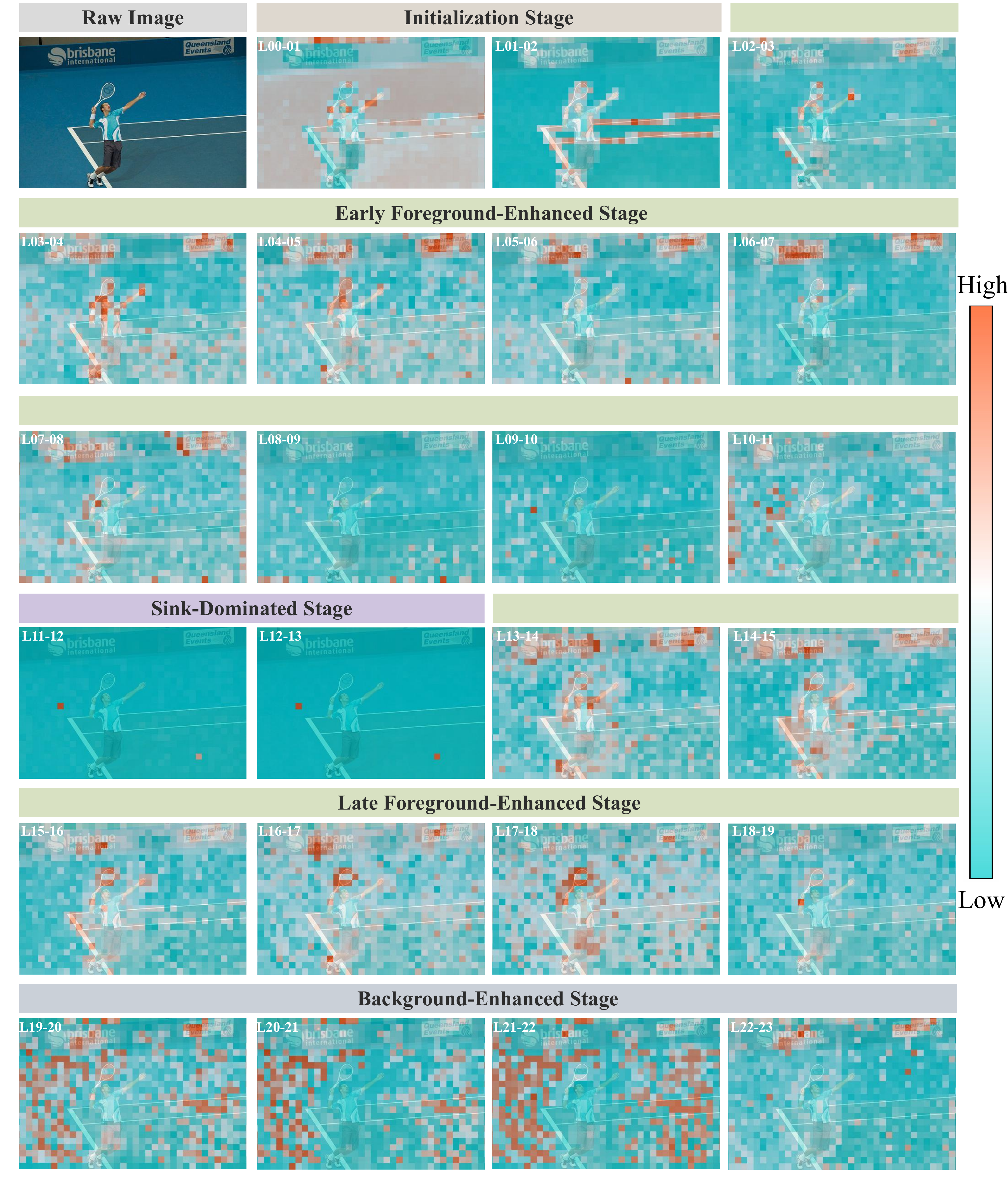}
\caption{Spatial maps of update magnitudes $v_i^{\ell}$ across LLaVA-NeXT vision encoder layers for the tennis scene.}
\label{fig:app-tennis-next}
\end{figure}

\begin{figure}[p]
\centering
\includegraphics[width=0.86\linewidth]{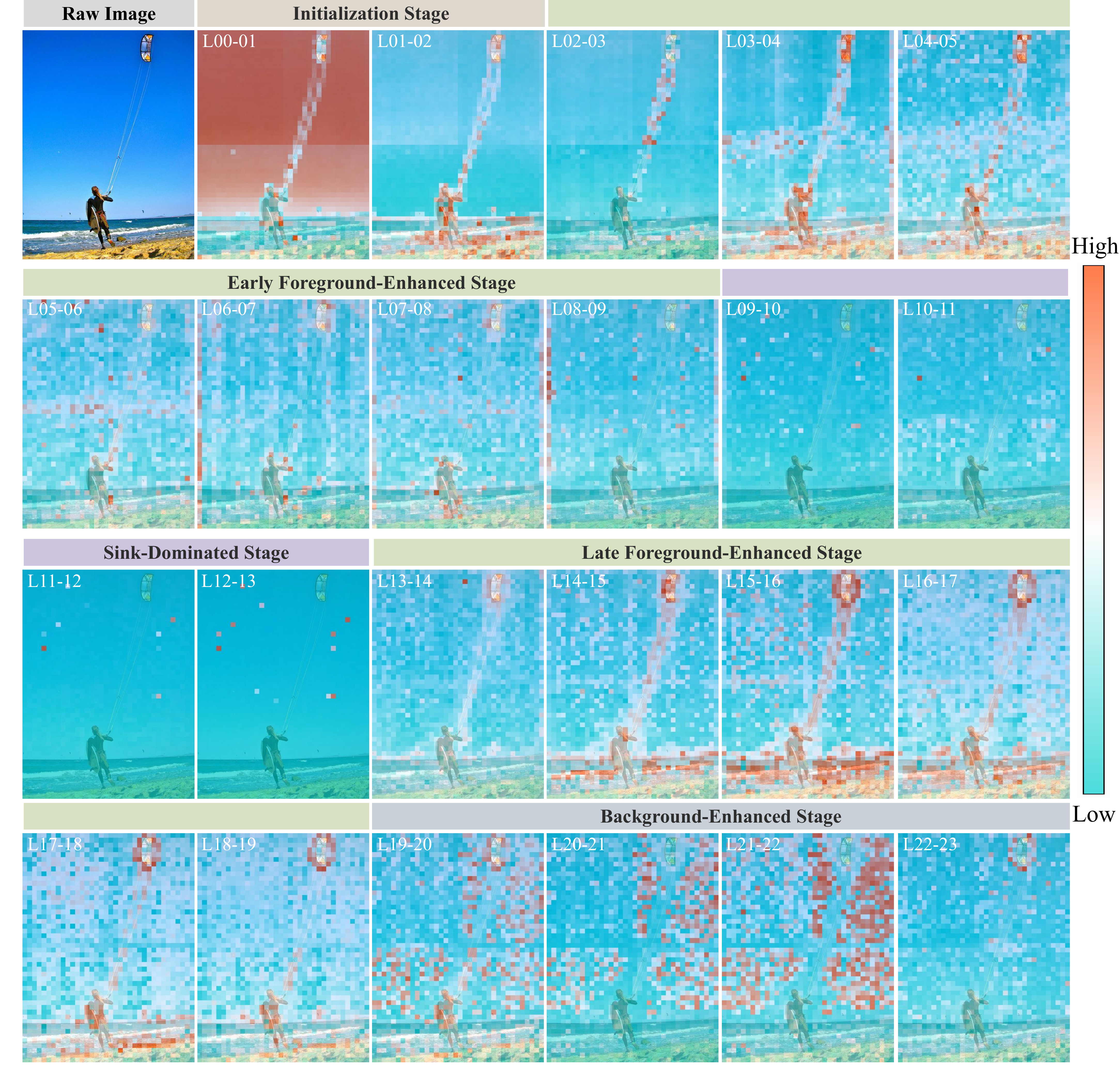}
\caption{Spatial maps of update magnitudes $v_i^{\ell}$ across LLaVA-NeXT vision encoder layers for the kite scene.}
\label{fig:app-kite-next}

\vspace{8pt}

\includegraphics[width=0.86\linewidth]{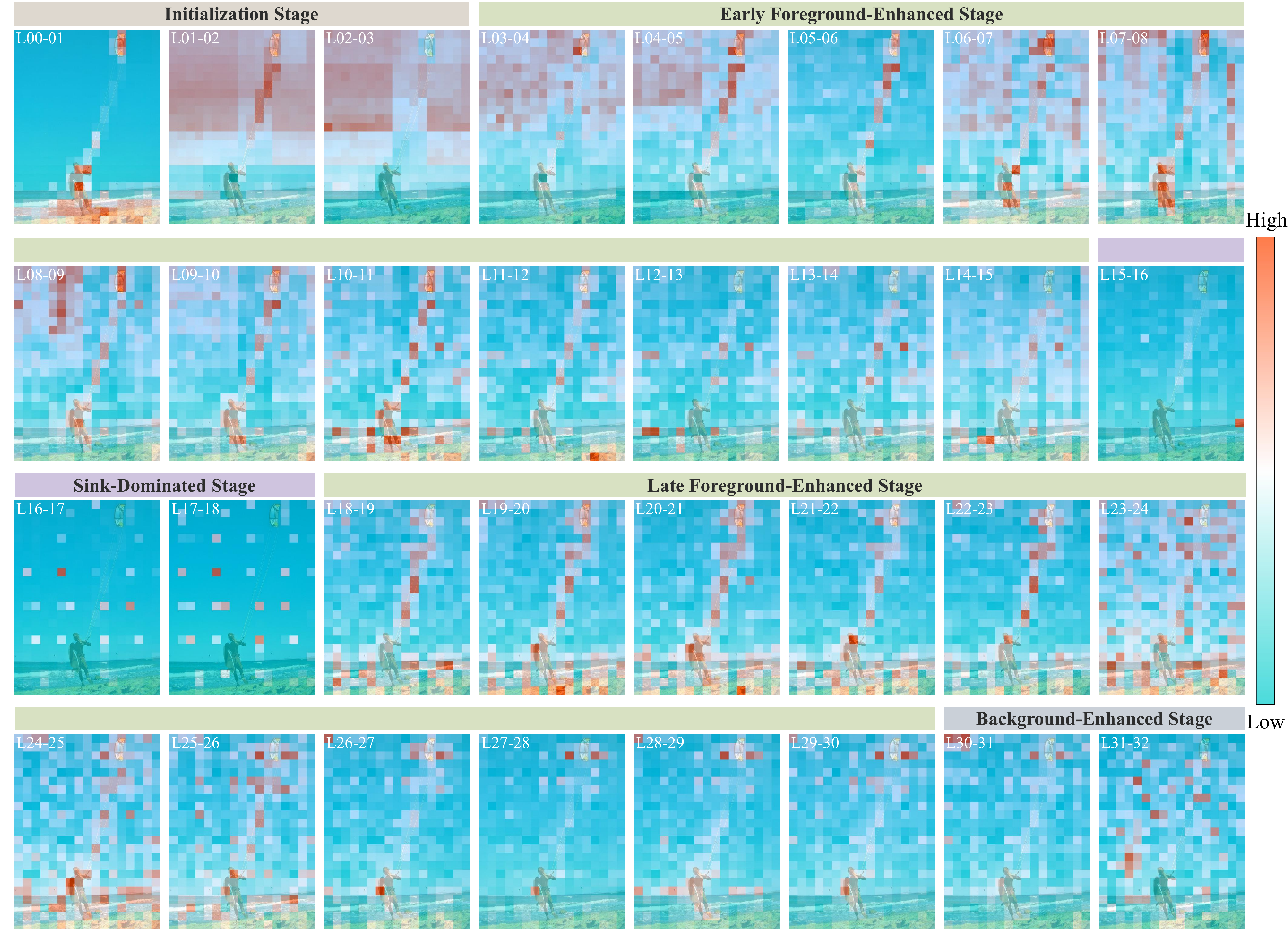}
\caption{Spatial maps of update magnitudes $v_i^{\ell}$ across Qwen2.5-VL vision encoder layers for the kite scene.}
\label{fig:app-kite-qwen}
\end{figure}
    
\begin{figure}[tbp]
\centering
\includegraphics[width=0.96\linewidth]{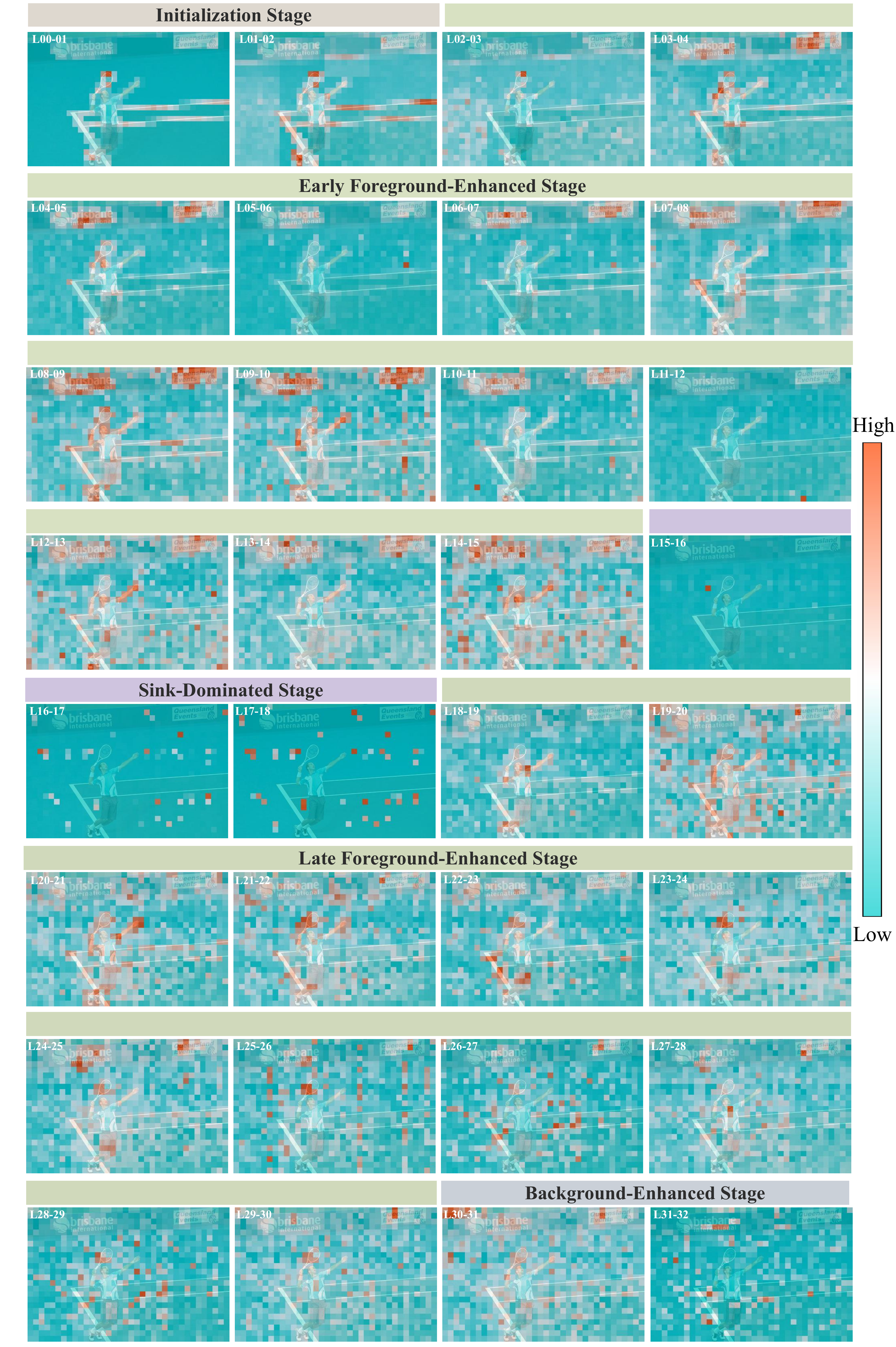}
\caption{Spatial maps of update magnitudes $v_i^{\ell}$ across Qwen2.5-VL vision encoder layers for the tennis scene.}
\label{fig:app-tennis-qwen}
\end{figure}

\endgroup
\FloatBarrier

\subsection{Update Direction Analysis}
\label{app:direction-analysis}
We compare update directions with encoder output features by semantic token retrieval on the COCO 2017 validation set~\citep{lin2014microsoft}, using COCO-Stuff labels~\citep{caesar2018coco}.
For each anchor, other labeled tokens in the same image are ranked by cosine similarity, and retrieval of the anchor class is scored.

\noindent
\begin{minipage}{\linewidth}
\centering
\captionsetup{skip=4pt,font=small}
\begin{minipage}[t]{0.48\linewidth}
\centering
\includegraphics[width=0.82\linewidth]{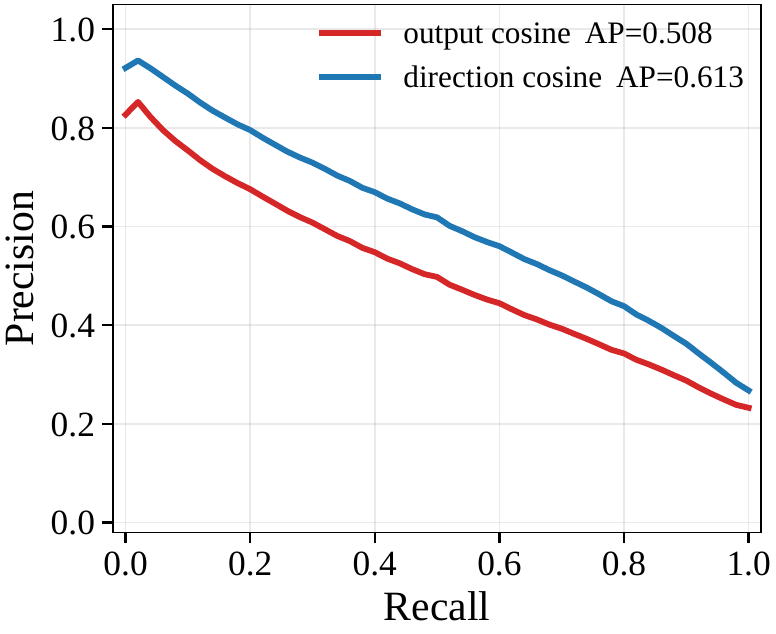}\\[1pt]
{\small (a) LLaVA-1.5}
\end{minipage}\hfill
\begin{minipage}[t]{0.48\linewidth}
\centering
\includegraphics[width=0.82\linewidth]{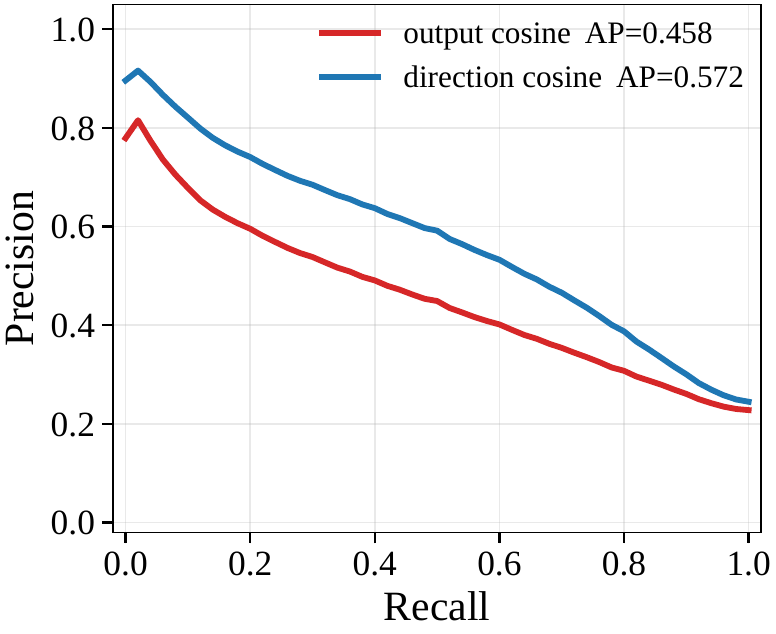}\\[1pt]
{\small (b) Qwen2.5-VL}
\end{minipage}
\captionof{figure}{Precision versus recall for semantic token retrieval with interior anchors on (a) LLaVA-1.5 and (b) Qwen2.5-VL.
The AP values shown in the legend are averaged over anchors.}
\label{fig:app-direction-pr}

\vspace{4pt}

\begin{minipage}[b]{0.245\linewidth}
\centering
\includegraphics[height=0.61\linewidth,trim=5.04bp 5.04bp 5.04bp 5.04bp,clip]{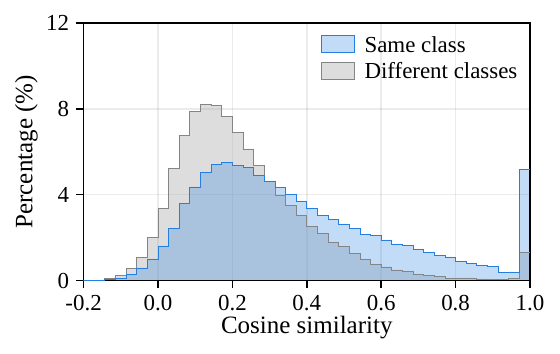}\\[1pt]
{\small (a) LLaVA-1.5\\ output features}
\end{minipage}\hfill%
\begin{minipage}[b]{0.245\linewidth}
\centering
\includegraphics[height=0.61\linewidth,trim=5.04bp 5.04bp 5.04bp 5.04bp,clip]{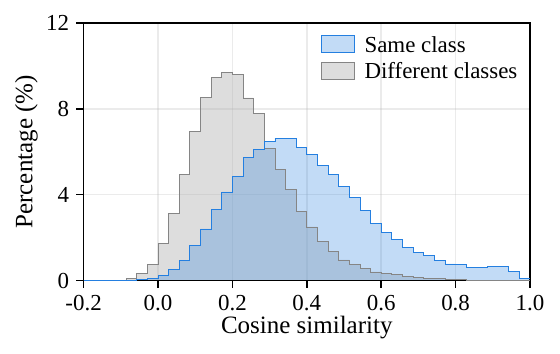}\\[1pt]
{\small (b) LLaVA-1.5\\ update directions}
\end{minipage}\hfill%
\begin{minipage}[b]{0.245\linewidth}
\centering
\includegraphics[height=0.61\linewidth]{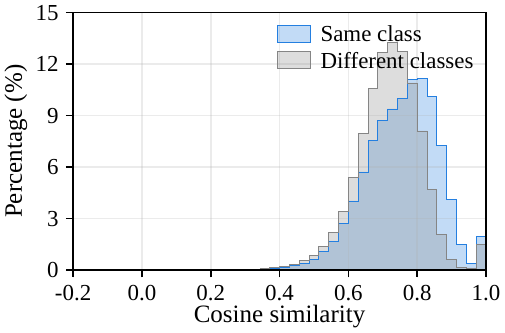}\\[1pt]
{\small (c) Qwen2.5-VL\\ output features}
\end{minipage}\hfill%
\begin{minipage}[b]{0.245\linewidth}
\centering
\includegraphics[height=0.61\linewidth]{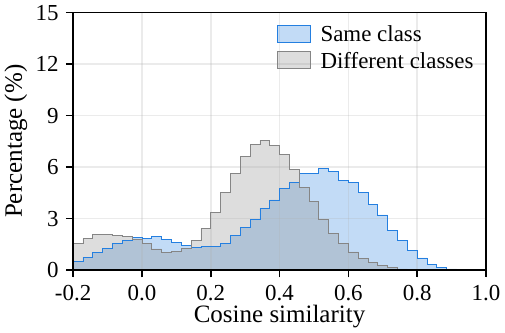}\\[1pt]
{\small (d) Qwen2.5-VL\\ update directions}
\end{minipage}
\captionof{figure}{Cosine similarity distributions for same-class (blue) and different-class (gray) pairs on LLaVA-1.5 (a, b) and Qwen2.5-VL (c, d).
Within each model, the left panel uses output features and the right panel uses update directions.
Panels (a) and (b) correspond to Figure~\ref{fig:score-distributions}.}
\label{fig:app-qwen-score-distributions}
\end{minipage}

\paragraph{Protocol.}
Images are encoded at \(336\times336\) with the CLIP-ViT-L/14 encoder in LLaVA-1.5 (\(24\times24\) tokens)~\citep{liu2024improved,radford2021learning}.
A token is assigned to class \(c\) at \(\ge 50\%\) patch coverage and marked interior at \(\ge 70\%\).
We keep image--class pairs with at least eight labeled tokens and select as the anchor the interior token that maximizes boundary distance times coverage, or the highest-coverage labeled token if none is interior.
Same-class tokens are positives, different-class tokens are negatives, and unlabeled tokens are excluded.
This yields \(25{,}158\) anchors on \(4{,}952\) images for LLaVA-1.5 and \(23{,}855\) anchors on \(4{,}949\) images for Qwen2.5-VL~\citep{bai2025qwen25vltechnicalreport}.

\paragraph{Features and metrics.}
Let \(h_i^{o}\) denote the encoder output feature of token \(i\): the final-layer feature for Qwen2.5-VL and the penultimate-layer feature for the CLIP ViT-L/14 encoder in LLaVA-1.5. We compare \(h_i^{o}\) with unit update directions \(d_i\) (Equation~\ref{eq:delta-h}; \(\ell_0=2\), \(\ell_d=L-1\)) using cosine similarity and the same candidate tokens.
Average precision (AP) is computed per anchor and then averaged over anchors, images, and classes; the image mean is the primary metric, with standard deviations over the same units.

\begin{figure}[tbp]
\centering
\captionsetup{hypcap=false,skip=4pt,font=small}

\begingroup
\captionof{table}{Semantic token retrieval AP on the COCO 2017 validation set for LLaVA-1.5.
Entries report mean AP over anchors, images, or classes; standard deviations are also reported for image and class means.
The mean over images is the primary metric; $\Delta$ is direction minus output.
Values are rounded to three decimal places.}
\label{tab:app-direction-ap}
\endgroup

\small
\setlength{\tabcolsep}{2.8pt}
\begin{tabular}{lccc}
\toprule[0.8pt]
Averaging & Output (AP $\uparrow$) & Direction (AP $\uparrow$) & $\Delta$ \\
\midrule[0.4pt]
Mean over anchors & $0.508$ & $0.613$ & 0.105 \\
Mean over images & $0.543{\pm}0.136$ & $0.649{\pm}0.130$ & 0.106 \\
Mean over classes & $0.532{\pm}0.108$ & $0.612{\pm}0.094$ & 0.081 \\
\bottomrule[0.8pt]
\end{tabular}

\vspace{8pt}
\includegraphics[width=\linewidth,height=0.75\textheight,keepaspectratio]{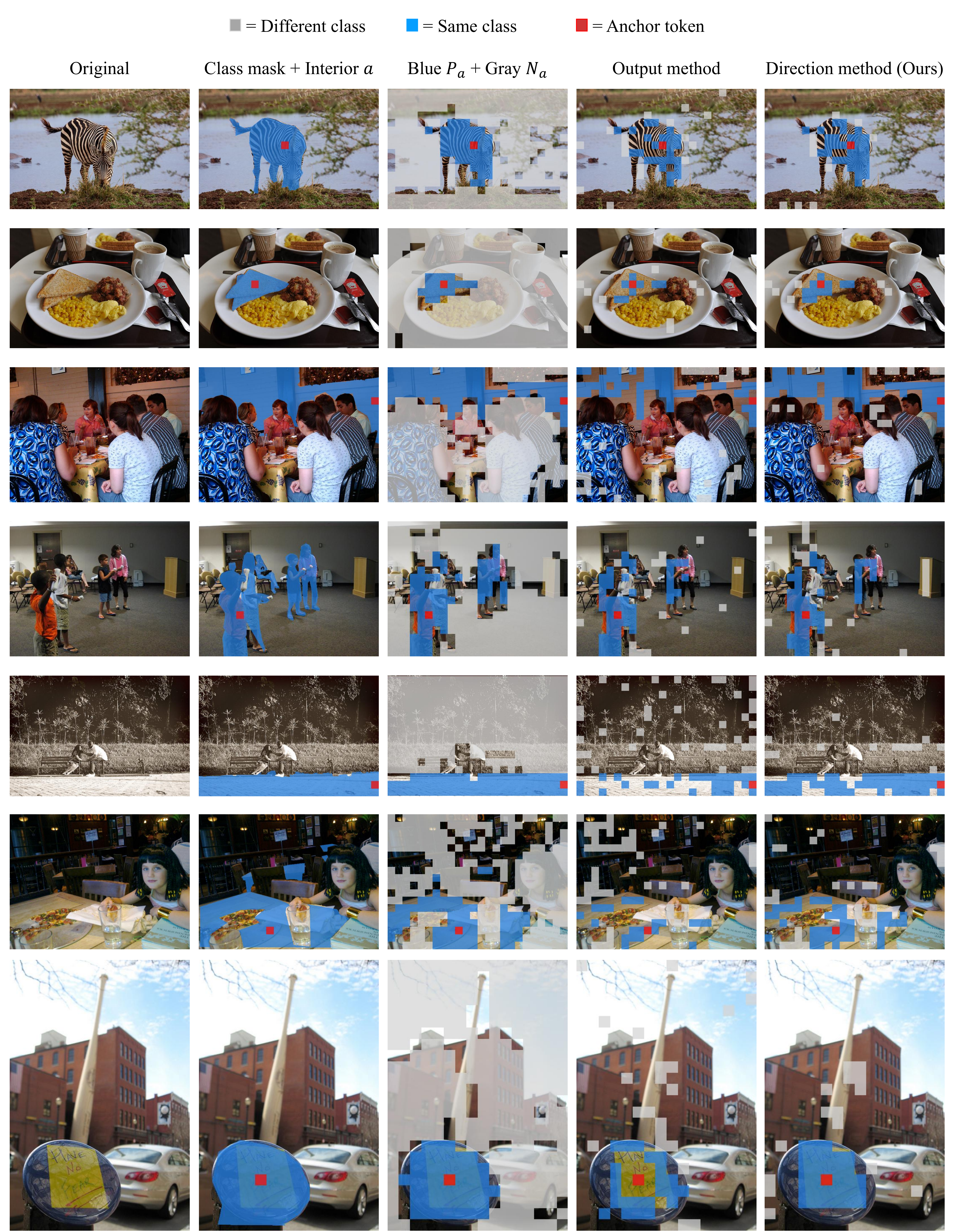}
\captionof{figure}{Semantic token retrieval examples.
From left to right: the input image; the class mask with anchor $a$; candidates $P_a$ (same class) and $N_a$ (different classes); and top-ranked tokens by cosine similarity to $a$ using output features and update directions.
The last two panels each show $k$ tokens, where $k$ is the number of interior tokens in the class mask.
``Interior $a$'' indicates at least $70\%$ coverage of the anchor patch by its class.
Candidates exclude the anchor and unlabeled tokens.
Red boxes mark $a$; blue and gray denote same-class and different-class tokens, respectively.
}
\label{fig:app-direction}
\end{figure}

\paragraph{Results.}
On LLaVA-1.5 (Table~\ref{tab:app-direction-ap}), update directions raise mean AP over images from \(0.543\) to \(0.649\), with gains on \(4{,}702/4{,}952\) images and \(162/171\) classes (Table~\ref{tab:app-direction-thing}); stuff classes improve more than thing classes (\(0.114\) vs.\ \(0.043\)).
Mean AP over anchors rises from \(0.508\) to \(0.613\) on LLaVA-1.5 and from \(0.458\) to \(0.572\) on Qwen2.5-VL (Figure~\ref{fig:app-direction-pr}).
Figure~\ref{fig:app-direction} shows fewer retrieved tokens from different classes when using update directions.

\begin{table*}[t]
\centering
\captionsetup{skip=4pt,font=small}
\caption{AP for each class on the COCO 2017 validation set: thing classes (top) and stuff classes (bottom).
$n$ is the number of anchors for the class, with one anchor per eligible image.
Out.\ and Dir.\ denote mean AP using cosine similarity of output features and update directions; $\Delta$ is Dir.\ minus Out.
Negative values of $\Delta$ are highlighted in red, indicating classes for which update directions retrieve same-class tokens less accurately than output features.}
\label{tab:app-direction-thing}
\label{tab:app-direction-stuff}
\scriptsize
\renewcommand{\arraystretch}{0.88}
{\footnotesize (a) Thing classes}\\[2pt]
\resizebox{\textwidth}{!}{%
\begin{tabular}{l@{\hspace{2.4pt}}r@{\hspace{2.4pt}}r@{\hspace{2.4pt}}r@{\hspace{2.4pt}}r@{\hspace{7pt}}l@{\hspace{2.4pt}}r@{\hspace{2.4pt}}r@{\hspace{2.4pt}}r@{\hspace{2.4pt}}r@{\hspace{7pt}}l@{\hspace{2.4pt}}r@{\hspace{2.4pt}}r@{\hspace{2.4pt}}r@{\hspace{2.4pt}}r}
\toprule[0.8pt]
Class & $n$ & Out. & Dir. & $\Delta$ & Class & $n$ & Out. & Dir. & $\Delta$ & Class & $n$ & Out. & Dir. & $\Delta$ \\
\midrule[0.4pt]
airplane & 72 & 0.555 & 0.610 & +0.055 & dining table & 405 & 0.536 & 0.619 & +0.083 & sandwich & 61 & 0.681 & 0.718 & +0.038 \\
apple & 35 & 0.615 & 0.650 & +0.035 & dog & 115 & 0.702 & 0.746 & +0.044 & scissors & 14 & 0.574 & 0.593 & +0.019 \\
backpack & 30 & 0.694 & 0.718 & +0.025 & donut & 40 & 0.738 & 0.745 & +0.008 & sheep & 52 & 0.656 & 0.701 & +0.045 \\
banana & 63 & 0.652 & 0.732 & +0.080 & elephant & 86 & 0.735 & 0.795 & +0.060 & sink & 87 & 0.472 & 0.539 & +0.067 \\
baseball bat & 7 & 0.618 & 0.605 & \textcolor{red}{-0.014} & fire hydrant & 49 & 0.620 & 0.679 & +0.059 & skateboard & 32 & 0.584 & 0.610 & +0.025 \\
baseball glove & 7 & 0.762 & 0.777 & +0.014 & fork & 27 & 0.511 & 0.533 & +0.022 & skis & 10 & 0.352 & 0.368 & +0.017 \\
bear & 44 & 0.664 & 0.739 & +0.075 & frisbee & 14 & 0.548 & 0.658 & +0.110 & snowboard & 13 & 0.343 & 0.445 & +0.102 \\
bed & 139 & 0.625 & 0.692 & +0.067 & giraffe & 94 & 0.677 & 0.720 & +0.043 & spoon & 19 & 0.394 & 0.417 & +0.023 \\
bench & 125 & 0.481 & 0.556 & +0.075 & hair drier & 3 & 0.601 & 0.670 & +0.069 & sports ball & 6 & 0.676 & 0.739 & +0.063 \\
bicycle & 61 & 0.567 & 0.578 & +0.012 & handbag & 47 & 0.528 & 0.566 & +0.038 & stop sign & 32 & 0.793 & 0.899 & +0.107 \\
bird & 56 & 0.712 & 0.742 & +0.030 & horse & 99 & 0.627 & 0.711 & +0.084 & suitcase & 59 & 0.588 & 0.680 & +0.093 \\
boat & 77 & 0.605 & 0.633 & +0.028 & hot dog & 31 & 0.688 & 0.718 & +0.030 & surfboard & 46 & 0.493 & 0.609 & +0.116 \\
book & 74 & 0.568 & 0.612 & +0.044 & keyboard & 70 & 0.652 & 0.715 & +0.063 & teddy bear & 74 & 0.674 & 0.705 & +0.031 \\
bottle & 96 & 0.618 & 0.587 & \textcolor{red}{-0.030} & kite & 30 & 0.648 & 0.698 & +0.050 & tennis racket & 35 & 0.475 & 0.502 & +0.027 \\
bowl & 148 & 0.561 & 0.595 & +0.035 & knife & 15 & 0.571 & 0.537 & \textcolor{red}{-0.034} & tie & 9 & 0.502 & 0.562 & +0.060 \\
broccoli & 42 & 0.604 & 0.666 & +0.062 & laptop & 127 & 0.540 & 0.596 & +0.057 & toaster & 3 & 0.734 & 0.735 & +0.002 \\
bus & 151 & 0.630 & 0.621 & \textcolor{red}{-0.010} & microwave & 24 & 0.535 & 0.565 & +0.030 & toilet & 123 & 0.485 & 0.557 & +0.072 \\
cake & 85 & 0.634 & 0.662 & +0.027 & motorcycle & 117 & 0.723 & 0.728 & +0.005 & toothbrush & 4 & 0.598 & 0.584 & \textcolor{red}{-0.014} \\
car & 202 & 0.668 & 0.658 & \textcolor{red}{-0.010} & mouse & 13 & 0.638 & 0.697 & +0.060 & traffic light & 33 & 0.595 & 0.639 & +0.044 \\
carrot & 41 & 0.695 & 0.761 & +0.065 & orange & 48 & 0.619 & 0.716 & +0.097 & train & 154 & 0.663 & 0.687 & +0.024 \\
cat & 152 & 0.701 & 0.773 & +0.072 & oven & 94 & 0.539 & 0.572 & +0.033 & truck & 138 & 0.656 & 0.663 & +0.007 \\
cell phone & 33 & 0.650 & 0.645 & \textcolor{red}{-0.005} & parking meter & 26 & 0.552 & 0.631 & +0.080 & tv & 149 & 0.607 & 0.646 & +0.038 \\
chair & 331 & 0.451 & 0.509 & +0.059 & person & 2037 & 0.604 & 0.626 & +0.021 & umbrella & 116 & 0.605 & 0.712 & +0.108 \\
clock & 52 & 0.625 & 0.656 & +0.032 & pizza & 105 & 0.710 & 0.752 & +0.042 & vase & 43 & 0.567 & 0.579 & +0.013 \\
couch & 169 & 0.522 & 0.613 & +0.091 & potted plant & 79 & 0.568 & 0.620 & +0.052 & wine glass & 36 & 0.412 & 0.425 & +0.012 \\
cow & 69 & 0.671 & 0.741 & +0.070 & refrigerator & 91 & 0.464 & 0.554 & +0.089 & zebra & 78 & 0.745 & 0.791 & +0.046 \\
cup & 139 & 0.576 & 0.588 & +0.012 & remote & 17 & 0.663 & 0.652 & \textcolor{red}{-0.012} &  &  &  &  &  \\
\bottomrule[0.8pt]
\end{tabular}%
}

\vspace{6pt}
{\footnotesize (b) Stuff classes}\\[2pt]
\resizebox{\textwidth}{!}{%
\begin{tabular}{l@{\hspace{2.4pt}}r@{\hspace{2.4pt}}r@{\hspace{2.4pt}}r@{\hspace{2.4pt}}r@{\hspace{7pt}}l@{\hspace{2.4pt}}r@{\hspace{2.4pt}}r@{\hspace{2.4pt}}r@{\hspace{2.4pt}}r@{\hspace{7pt}}l@{\hspace{2.4pt}}r@{\hspace{2.4pt}}r@{\hspace{2.4pt}}r@{\hspace{2.4pt}}r}
\toprule[0.8pt]
Class & $n$ & Out. & Dir. & $\Delta$ & Class & $n$ & Out. & Dir. & $\Delta$ & Class & $n$ & Out. & Dir. & $\Delta$ \\
\midrule[0.4pt]
banner & 60 & 0.563 & 0.648 & +0.085 & furniture-other & 539 & 0.402 & 0.475 & +0.073 & salad & 11 & 0.713 & 0.702 & \textcolor{red}{-0.010} \\
blanket & 46 & 0.451 & 0.559 & +0.108 & grass & 734 & 0.503 & 0.682 & +0.179 & sand & 183 & 0.498 & 0.678 & +0.180 \\
branch & 46 & 0.468 & 0.555 & +0.087 & gravel & 84 & 0.358 & 0.485 & +0.127 & sea & 261 & 0.671 & 0.744 & +0.073 \\
bridge & 54 & 0.474 & 0.551 & +0.077 & ground-other & 113 & 0.422 & 0.589 & +0.167 & shelf & 155 & 0.367 & 0.447 & +0.080 \\
building-other & 744 & 0.578 & 0.619 & +0.041 & hill & 71 & 0.508 & 0.643 & +0.135 & sky-other & 1080 & 0.474 & 0.801 & +0.327 \\
bush & 274 & 0.451 & 0.564 & +0.113 & house & 185 & 0.539 & 0.581 & +0.042 & skyscraper & 59 & 0.577 & 0.638 & +0.061 \\
cabinet & 254 & 0.412 & 0.534 & +0.123 & leaves & 79 & 0.550 & 0.630 & +0.080 & snow & 174 & 0.651 & 0.795 & +0.144 \\
cage & 87 & 0.455 & 0.554 & +0.099 & light & 59 & 0.459 & 0.551 & +0.092 & solid-other & 19 & 0.476 & 0.570 & +0.094 \\
cardboard & 105 & 0.538 & 0.607 & +0.069 & mat & 15 & 0.364 & 0.534 & +0.170 & stairs & 40 & 0.450 & 0.546 & +0.096 \\
carpet & 128 & 0.390 & 0.552 & +0.162 & metal & 582 & 0.412 & 0.483 & +0.071 & stone & 34 & 0.507 & 0.590 & +0.082 \\
ceiling-other & 320 & 0.416 & 0.576 & +0.160 & mirror-stuff & 102 & 0.365 & 0.396 & +0.031 & straw & 48 & 0.443 & 0.577 & +0.134 \\
ceiling-tile & 9 & 0.430 & 0.626 & +0.196 & moss & 7 & 0.436 & 0.578 & +0.142 & structural-other & 90 & 0.433 & 0.518 & +0.085 \\
cloth & 30 & 0.503 & 0.578 & +0.076 & mountain & 143 & 0.511 & 0.621 & +0.110 & table & 383 & 0.416 & 0.533 & +0.117 \\
clothes & 191 & 0.445 & 0.502 & +0.056 & mud & 11 & 0.522 & 0.635 & +0.113 & tent & 37 & 0.476 & 0.538 & +0.062 \\
clouds & 322 & 0.537 & 0.799 & +0.262 & napkin & 32 & 0.534 & 0.608 & +0.074 & textile-other & 200 & 0.491 & 0.578 & +0.087 \\
counter & 95 & 0.332 & 0.427 & +0.095 & net & 39 & 0.373 & 0.453 & +0.080 & towel & 39 & 0.592 & 0.610 & +0.017 \\
cupboard & 16 & 0.357 & 0.533 & +0.176 & paper & 196 & 0.501 & 0.560 & +0.060 & tree & 1277 & 0.554 & 0.666 & +0.113 \\
curtain & 170 & 0.491 & 0.600 & +0.110 & pavement & 641 & 0.402 & 0.548 & +0.146 & vegetable & 49 & 0.561 & 0.624 & +0.062 \\
desk-stuff & 114 & 0.414 & 0.524 & +0.110 & pillow & 13 & 0.569 & 0.717 & +0.149 & wall-brick & 137 & 0.463 & 0.584 & +0.121 \\
dirt & 321 & 0.425 & 0.573 & +0.148 & plant-other & 166 & 0.488 & 0.586 & +0.098 & wall-concrete & 1086 & 0.437 & 0.616 & +0.179 \\
door-stuff & 246 & 0.374 & 0.504 & +0.130 & plastic & 265 & 0.468 & 0.512 & +0.044 & wall-other & 451 & 0.414 & 0.558 & +0.144 \\
fence & 250 & 0.482 & 0.573 & +0.091 & platform & 75 & 0.390 & 0.548 & +0.158 & wall-panel & 127 & 0.419 & 0.586 & +0.167 \\
floor-marble & 41 & 0.356 & 0.521 & +0.165 & playingfield & 217 & 0.608 & 0.809 & +0.201 & wall-stone & 59 & 0.520 & 0.599 & +0.079 \\
floor-other & 253 & 0.345 & 0.522 & +0.176 & railing & 31 & 0.460 & 0.530 & +0.070 & wall-tile & 194 & 0.460 & 0.566 & +0.106 \\
floor-stone & 51 & 0.430 & 0.585 & +0.155 & railroad & 100 & 0.414 & 0.491 & +0.077 & wall-wood & 225 & 0.406 & 0.551 & +0.144 \\
floor-tile & 258 & 0.395 & 0.521 & +0.126 & river & 79 & 0.570 & 0.708 & +0.138 & water-other & 65 & 0.518 & 0.622 & +0.104 \\
floor-wood & 193 & 0.422 & 0.579 & +0.157 & road & 538 & 0.416 & 0.584 & +0.168 & waterdrops & 3 & 0.237 & 0.270 & +0.033 \\
flower & 50 & 0.575 & 0.606 & +0.031 & rock & 81 & 0.547 & 0.635 & +0.087 & window-blind & 73 & 0.462 & 0.625 & +0.163 \\
fog & 102 & 0.452 & 0.784 & +0.332 & roof & 76 & 0.468 & 0.580 & +0.112 & window-other & 331 & 0.499 & 0.600 & +0.101 \\
food-other & 130 & 0.587 & 0.608 & +0.021 & rug & 66 & 0.460 & 0.591 & +0.131 & wood & 97 & 0.460 & 0.553 & +0.093 \\
fruit & 41 & 0.502 & 0.541 & +0.039 &  &  &  &  &  &  &  &  &  &  \\
\bottomrule[0.8pt]
\end{tabular}%
}
\end{table*}

\clearpage
\section{Additional Experimental Results}
\label{app:extra-exp}
This section describes sink detection, compression settings, and evaluation metrics, followed by results on larger models and component ablations.
\subsection{Sink Detection Configuration}
\label{app:sink-detection}
For each encoder, we locate the peak of the image-averaged max-to-median update magnitude ratio:
\begin{equation}
    \bar r^\ell=\frac{1}{|\mathcal D|}\sum_{I\in\mathcal D}
    \frac{\max_i v_i^\ell(I)}{\operatorname{median}_i v_i^\ell(I)},
    \qquad
    \ell_{\mathrm{peak}}=\arg\max_\ell\bar r^\ell,
    \label{eq:sink-peak-layer}
\end{equation}
where $\mathcal D$ contains the analysis images and $v_i^\ell(I)=\lVert h_i^{\ell+1}(I)-h_i^\ell(I)\rVert$.
We set $\ell_\star:=\ell_{\mathrm{peak}}$ and inspect $h_i^{\ell_{\mathrm{peak}}+1}$, the output of the peak layer.

Motivated by the observation that sparse activation coordinates can drive high-norm outlier tokens~\citep{jiang2025vision}, we examine the activation entropy of the ten highest-norm tokens at this state and identify a diagnostic coordinate $d_s$ from the concentrated activations of low-entropy candidates.
We use $d_s=650$ for CLIP-ViT-L and $d_s=849$ for the Qwen2.5-VL vision encoder, with an empirical threshold $\tau=50$.
At inference, tokens with absolute activation above $\tau$ are excluded:
\begin{equation}
    \mathcal V=\left\{i:\left|\left(h_i^{\ell_\star+1}\right)_{d_s}\right|\leq\tau\right\}.
    \label{eq:sink-detection}
\end{equation}
These detection settings remain fixed across datasets, images, and token budgets; norm ranking and entropy analysis are performed only when configuring the encoder.

\subsection{Experimental Settings and Evaluation Metrics}
\label{app:exp-details}
\paragraph{Compression settings.}
We form $\Delta h_i$ as in Equation~\ref{eq:delta-h}, normalize it to unit length, and group the resulting directions with spherical $k$-means~\citep{hornik2012spherical}.
Centroids are initialized by deterministic farthest-point sampling as in Equation~\ref{eq:farthest-sampling}, with the initialization seed fixed to $s=0$ in all experiments.
Unless otherwise stated, each grouping run stops when the assignment labels remain unchanged,
the relative change in grouping loss is below $10^{-5}$, or $10$ iterations have been completed.
We use $K=20$ for all models in the main experiments and evaluate the effect of the group count in Appendix~\ref{app:ablation-k}.
The default grouping endpoints are $\ell_0=2$ and $\ell_d=L-1$.
The saliency windows are $14\to19$ for CLIP-ViT-L and $19\to24$ for the Qwen2.5-VL vision encoder, following the layer convention in Section~\ref{sec:foreground-window-saliency-estimation}.
For a given encoder, these settings are held fixed across datasets, images, and token budgets.
Unless otherwise stated, all experiments are run on a single NVIDIA RTX PRO 6000-96\,GB GPU.

\paragraph{Qwen2.5-VL adaptation and budget convention.}
\label{app:qwen-alignment}
For Qwen2.5-VL~\citep{bai2025qwen25vltechnicalreport}, update directions and visual saliency scores are computed separately on the original encoder patches before aggregation.
These signals are then aggregated over each set of four patches combined by the spatial merger and aligned with the corresponding visual token passed to the LLM.
If any of these four patches is identified as a sink, the corresponding visual token is excluded from the candidate set.
The visual token count $N$ and target budget $B$ refer to tokens after merging.
The reported retention ratio $B/N$ uses the visual token count before sink removal as its denominator.

\paragraph{Mini-Gemini adaptation.}
For Mini-Gemini~\citep{li2025mini}, visual tokens come from the low-resolution CLIP branch and follow the same settings as in LLaVA-1.5.

\paragraph{Evaluation metrics.}
We use LMMs-Eval~\citep{zhang2025lmms} for downstream evaluation.
RelAcc.\ is the mean relative benchmark score: each score is divided by the corresponding uncompressed score, and the resulting ratios are averaged across the benchmarks in the table and expressed as a percentage.

\subsection{Evaluation on Larger Models}
\label{app:larger-models}
We extend the evaluation from the 7B models in the main text to LLaVA-1.5-13B~\citep{liu2024improved}, LLaVA-NeXT-13B~\citep{liu2024llavanext}, and Qwen2.5-VL-32B~\citep{bai2025qwen25vltechnicalreport}.
These larger models share the vision encoder architectures of their 7B counterparts, allowing us to assess the method at increased scale within the same model families.

LLaVA-1.5-13B encodes each image into 576 visual tokens.
Following VisionZip~\citep{yang2025visionzip} and PruneSID~\citep{fang2026prune}, we retain 192, 128, and 64 tokens.
In Table~\ref{tab:llava15-13b}, \ours{} achieves the highest RelAcc.\ at all three budgets, with scores of 98.3\%, 97.4\%, and 96.3\%.
The gains over the stronger baseline at each budget are 0.6, 0.1, and 1.3 percentage points, respectively.

LLaVA-NeXT-13B produces up to 2,880 visual tokens; we retain 640, 320, and 160 tokens, matching the budgets used for the 7B model.
Table~\ref{tab:llava-next-13b} compares \ours{} with PruneSID.
Our method exceeds PruneSID by 1.0 and 0.3 percentage points at 640 and 160 tokens and matches it at 94.0\% at 320 tokens.

For Qwen2.5-VL-32B, we report retention ratios because the visual token count depends on the input, consistent with Table~\ref{tab:qwen25vl}.
As reported in Table~\ref{tab:qwen25vl-32b}, \ours{} achieves RelAcc.\ of 96.6\%, 94.7\%, and 89.8\% at retention ratios of 33.3\%, 22.2\%, and 11.1\%, respectively.
Its advantage over the stronger baseline at each ratio is 0.4, 2.3, and 2.8 percentage points, increasing as retention decreases.
Together, these results support the generalization of \ours{} to larger models across different multimodal architectures.

\begingroup
\captionsetup{skip=4pt}
\begin{table}[ht]
\centering
\caption{Performance comparison on LLaVA-1.5-13B.
RelAcc.\ averages the relative scores across the nine listed benchmarks.}
\label{tab:llava15-13b}
\setlength{\tabcolsep}{3.2pt}
\renewcommand{\arraystretch}{0.9}
\resizebox{0.82\linewidth}{!}{%
\begin{tabular}{lcccccccccc}
\toprule[0.8pt]
Method & GQA & MMB & MME & POPE & SQA & VQA$^{\text{v2}}$ & VQA$^{\text{Text}}$ & SEED$^{\text{I}}$ & VizWiz & RelAcc. \\
\midrule[0.4pt]
\multicolumn{11}{l}{\textcolor{black!60}{Uncompressed baseline (100\%)}} \\
Vanilla & 63.2 & 67.7 & 1818 & 85.9 & 72.8 & 80.0 & 61.3 & 66.9 & 56.6 & 100\% \\
\specialrule{0.3pt}{3pt}{3pt}
\multicolumn{11}{l}{\textcolor{black!60}{Token retention: 33.3\%}} \\
VisionZip & 59.6 & 65.9 & 1770 & 86.4 & 72.8 & 78.0 & 58.6 & 65.2 & 54.9 & 97.5\% \\
PruneSID & 59.6 & 65.9 & 1770 & 86.4 & 72.8 & 78.0 & 58.6 & 65.2 & 56.0 & 97.7\% \\
\rowcolor{tableblue!60}
Ours & 60.1 & 67.4 & 1801 & 86.1 & 73.7 & 76.8 & 59.4 & 65.3 & 56.0 & \textbf{98.3\%} \\
\specialrule{0.3pt}{3pt}{3pt}
\multicolumn{11}{l}{\textcolor{black!60}{Token retention: 22.2\%}} \\
VisionZip & 57.9 & 66.7 & 1743 & 85.2 & 74.0 & 76.8 & 58.7 & 63.8 & 55.0 & 96.8\% \\
PruneSID & 58.9 & 65.5 & 1811 & 85.9 & 73.1 & 76.7 & 57.5 & 64.1 & 56.8 & 97.3\% \\
\rowcolor{tableblue!60}
Ours & 59.2 & 65.9 & 1783 & 86.6 & 73.3 & 75.9 & 58.8 & 64.2 & 56.1 & \textbf{97.4\%} \\
\specialrule{0.3pt}{3pt}{3pt}
\multicolumn{11}{l}{\textcolor{black!60}{Token retention: 11.1\%}} \\
VisionZip & 56.2 & 64.9 & 1676 & 76.0 & 74.4 & 73.7 & 57.4 & 60.4 & 55.9 & 93.6\% \\
PruneSID & 57.8 & 63.8 & 1711 & 82.0 & 71.8 & 75.2 & 56.3 & 62.8 & 57.3 & 95.0\% \\
\rowcolor{tableblue!60}
Ours & 58.4 & 64.6 & 1792 & 85.3 & 72.4 & 74.1 & 57.5 & 62.4 & 57.5 & \textbf{96.3\%} \\
\bottomrule[0.8pt]
\end{tabular}%
}
\end{table}

\begin{table}[ht]
\centering
\caption{Performance comparison on LLaVA-NeXT-13B.}
\label{tab:llava-next-13b}
\setlength{\tabcolsep}{3.2pt}
\renewcommand{\arraystretch}{0.9}
\resizebox{0.82\linewidth}{!}{%
\begin{tabular}{lcccccccccc}
\toprule[0.8pt]
Method & GQA & MMB & MME & POPE & SQA & VQA$^{\text{v2}}$ & VQA$^{\text{Text}}$ & SEED$^{\text{I}}$ & VizWiz & RelAcc. \\
\midrule[0.4pt]
\multicolumn{11}{l}{\textcolor{black!60}{Uncompressed baseline (100\%)}} \\
Vanilla & 65.4 & 70.0 & 1858 & 86.2 & 73.5 & 81.8 & 64.3 & 71.9 & 64.0 & 100\% \\
\specialrule{0.3pt}{3pt}{3pt}
\multicolumn{11}{l}{\textcolor{black!60}{Token retention: 22.2\%}} \\
PruneSID & 62.4 & 67.0 & 1817 & 85.6 & 70.1 & 79.1 & 60.2 & 68.5 & 60.2 & 95.9\% \\
\rowcolor{tableblue!60}
Ours & 63.6 & 66.6 & 1820 & 87.0 & 72.8 & 78.6 & 62.0 & 68.5 & 60.1 & \textbf{96.9\%} \\
\specialrule{0.3pt}{3pt}{3pt}
\multicolumn{11}{l}{\textcolor{black!60}{Token retention: 11.1\%}} \\
PruneSID & 61.5 & 65.4 & 1810 & 82.7 & 70.3 & 76.9 & 58.5 & 66.8 & 58.5 & \textbf{94.0\%} \\
\rowcolor{tableblue!60}
Ours & 61.3 & 64.5 & 1753 & 86.9 & 73.1 & 75.6 & 60.1 & 64.9 & 57.0 & \textbf{94.0\%} \\
\specialrule{0.3pt}{3pt}{3pt}
\multicolumn{11}{l}{\textcolor{black!60}{Token retention: 5.6\%}} \\
PruneSID & 59.5 & 65.5 & 1715 & 77.8 & 69.1 & 73.6 & 56.7 & 64.1 & 56.7 & 90.8\% \\
\rowcolor{tableblue!60}
Ours & 58.5 & 63.3 & 1712 & 85.7 & 72.5 & 72.9 & 56.3 & 62.2 & 55.8 & \textbf{91.1\%} \\
\bottomrule[0.8pt]
\end{tabular}%
}
\end{table}

\begin{table}[ht]
\centering
\caption{Performance comparison on Qwen2.5-VL-32B.}
\label{tab:qwen25vl-32b}
\setlength{\tabcolsep}{6.4pt}
\renewcommand{\arraystretch}{0.90}
\resizebox{0.82\linewidth}{!}{%
\begin{tabular}{lcccccccc}
\toprule[0.8pt]
Method & GQA & MMB & MME & POPE & SQA & MMMU & VQA$^{\text{Text}}$ & RelAcc. \\
\midrule[0.4pt]
\multicolumn{9}{l}{\textcolor{black!60}{Uncompressed baseline (100\%)}} \\
Vanilla & 59.3 & 86.3 & 2428 & 84.3 & 91.5 & 61.1 & 76.5 & 100\% \\
\specialrule{0.3pt}{3pt}{3pt}
\multicolumn{9}{l}{\textcolor{black!60}{Token retention: 33.3\%}} \\
VisionZip & 57.3 & 85.1 & 2307 & 82.1 & 88.9 & 57.7 & 72.1 & 96.2\% \\
PruneSID & 57.0 & 81.6 & 2209 & 82.7 & 85.4 & 57.6 & 70.6 & 94.2\% \\
\rowcolor{tableblue!60}
Ours & 56.4 & 84.4 & 2396 & 81.6 & 89.1 & 58.3 & 72.9 & \textbf{96.6\%} \\
\specialrule{0.3pt}{3pt}{3pt}
\multicolumn{9}{l}{\textcolor{black!60}{Token retention: 22.2\%}} \\
VisionZip & 55.6 & 82.3 & 2173 & 79.8 & 87.0 & 55.3 & 67.0 & 92.4\% \\
PruneSID & 55.9 & 80.8 & 2158 & 82.1 & 83.6 & 55.8 & 66.9 & 92.0\% \\
\rowcolor{tableblue!60}
Ours & 55.2 & 81.4 & 2356 & 80.1 & 87.5 & 58.3 & 70.5 & \textbf{94.7\%} \\
\specialrule{0.3pt}{3pt}{3pt}
\multicolumn{9}{l}{\textcolor{black!60}{Token retention: 11.1\%}} \\
VisionZip & 51.8 & 77.4 & 1862 & 71.4 & 82.5 & 56.4 & 57.8 & 85.2\% \\
PruneSID & 52.6 & 73.9 & 2076 & 75.0 & 81.3 & 55.8 & 61.3 & 87.0\% \\
\rowcolor{tableblue!60}
Ours & 52.5 & 77.3 & 2175 & 78.3 & 82.5 & 56.1 & 65.7 & \textbf{89.8\%} \\
\bottomrule[0.8pt]
\end{tabular}%
}
\end{table}
\endgroup

\subsection{Grouping, Saliency Scoring, and Update Aggregation on LLaVA-1.5}
\label{app:llava-cluster}
\label{app:llava-update-accum}
\label{app:llava-saliency-score}
Table~\ref{tab:llava-cluster} compares three alternatives on LLaVA-1.5-7B, changing one component at a time.

\noindent\textbf{Grouping.}
Grouping by update directions $d_i$ improves RelAcc.\ over encoder output features $h_i^o$ by 0.7, 0.9, and 0.7 percentage points at the three retention ratios.
Together with the Qwen2.5-VL results in Table~\ref{tab:ablation-cluster}, these consistent downstream gains support update directions as a grouping representation.

\noindent\textbf{Saliency scoring.}
The visual saliency score $u_i$, computed from endpoint displacement, outperforms output-layer [CLS] attention by 1.4, 1.5, and 2.2 percentage points in RelAcc., supporting update magnitudes for visual saliency estimation.

\noindent\textbf{Update aggregation.}
Accumulated update magnitude matches or exceeds endpoint displacement by up to 0.3 points on LLaVA-1.5, while endpoint displacement performs better on Qwen2.5-VL (Table~\ref{tab:qwen-update-accum}).
Endpoint displacement computes each score from two endpoint states with a single difference and norm, avoiding calculations over every intermediate transition.
These results and the reduced computation and state-access requirements support endpoint displacement as the shared default.

\begin{table}[ht]
\centering
\captionsetup{skip=4pt}
\caption{Grouping, saliency scoring, and update aggregation on LLaVA-1.5-7B.
Ours uses unit update directions and endpoint displacement.
Output features replaces the grouping representation with $h_i^{o}$.
Attention replaces the visual saliency score $u_i$ with [CLS] attention from the output layer.
Accumulated replaces endpoint displacement with accumulated update magnitude.}
\label{tab:llava-cluster}
\label{tab:llava-update-accum}
\label{tab:llava-saliency-score}
\setlength{\tabcolsep}{2.2pt}
\renewcommand{\arraystretch}{0.86}
\resizebox{\linewidth}{!}{%
\begin{tabular}{lrrrrrrrrrrrrrrr}
\toprule[0.8pt]
\multirow{2}{*}{Method} & \multicolumn{5}{c}{\textcolor{black!60}{Retain 33.3\%}}
& \multicolumn{5}{c}{\textcolor{black!60}{Retain 22.2\%}}
& \multicolumn{5}{c}{\textcolor{black!60}{Retain 11.1\%}} \\
\cmidrule(lr){2-6}\cmidrule(lr){7-11}\cmidrule(lr){12-16}
& GQA & MME & POPE & SQA & RelAcc.
& GQA & MME & POPE & SQA & RelAcc.
& GQA & MME & POPE & SQA & RelAcc. \\
\midrule[0.4pt]
\rowcolor{tableblue!60}
Ours & 60.1 & 1806 & 86.9 & 68.7 & 98.5\%
& 58.9 & 1778 & 86.7 & 68.7 & \textbf{97.6\%}
& 57.7 & 1708 & 85.6 & 68.4 & 95.8\% \\
\specialrule{0.3pt}{3pt}{3pt}
\multicolumn{16}{l}{\textcolor{black!60}{Grouping}} \\
Output features & 60.0 & 1771 & 86.7 & 68.4 & 97.8\%
& 58.9 & 1712 & 86.6 & 68.7 & 96.7\%
& 57.8 & 1680 & 84.4 & 68.4 & 95.1\% \\
\specialrule{0.3pt}{3pt}{3pt}
\multicolumn{16}{l}{\textcolor{black!60}{Saliency scoring}} \\
Attention & 58.7 & 1795 & 84.3 & 68.7 & 97.1\%
& 57.9 & 1775 & 83.3 & 68.4 & 96.1\%
& 56.4 & 1675 & 81.2 & 68.7 & 93.6\% \\
\specialrule{0.3pt}{3pt}{3pt}
\multicolumn{16}{l}{\textcolor{black!60}{Update aggregation}} \\
Accumulated & 60.1 & 1800 & 87.0 & 69.1 & \textbf{98.6\%}
& 59.1 & 1770 & 87.0 & 68.4 & \textbf{97.6\%}
& 57.8 & 1719 & 86.5 & 68.0 & \textbf{96.1\%} \\
\bottomrule[0.8pt]
\end{tabular}%
}
\end{table}

\subsection{Grouping Endpoints}
We vary the early endpoint $\ell_0$ and the late endpoint $\ell_d$ used to compute update directions.
Table~\ref{tab:ablation-init} varies $\ell_0$ with $\ell_d=L-1$, while Table~\ref{tab:ablation-end} varies $\ell_d$ with $\ell_0=2$.
Here, $h^{L-1}$ is the output of Layer $L-2$.
Using the encoder input ($\ell_0=0$) or final output ($\ell_d=L$) reduces mean RelAcc.\ by 0.5 and 0.7 percentage points relative to the default pair $(2,L-1)$.
In contrast, varying $\ell_0$ within $\{1,2,3\}$ or $\ell_d$ within $\{L-3,L-2,L-1\}$ changes mean RelAcc.\ by at most 0.3 points.
These results support the default endpoints and show limited sensitivity to nearby choices.

\begingroup
\begin{table}[ht]
\centering
\captionsetup{skip=4pt}
\caption{Ablation of the early endpoint $\ell_0$ on Qwen2.5-VL-7B.
Avg.\ is the mean RelAcc.\ over the three retention ratios.
}
\label{tab:ablation-init}
\setlength{\tabcolsep}{1.6pt}
\renewcommand{\arraystretch}{0.86}
\resizebox{\linewidth}{!}{%
\begin{tabular}{cccccccccccccccccccc}
\toprule[0.8pt]
& \multicolumn{6}{c}{\textcolor{black!60}{Retain 33.3\%}}
& \multicolumn{6}{c}{\textcolor{black!60}{Retain 22.2\%}}
& \multicolumn{6}{c}{\textcolor{black!60}{Retain 11.1\%}} & \\
\cmidrule(lr){2-7}\cmidrule(lr){8-13}\cmidrule(lr){14-19}
\makebox[2.9em][c]{$\ell_0$} & GQA & MMB & MME & POPE & SQA & RelAcc.
& GQA & MMB & MME & POPE & SQA & RelAcc.
& GQA & MMB & MME & POPE & SQA & RelAcc. & Avg. \\
\midrule[0.4pt]
\makebox[2.9em][c]{$0$} & 58.1 & 81.7 & 2323 & 84.9 & 88.0 & 98.2\%
& 57.0 & 80.2 & 2300 & 83.7 & 87.1 & 96.8\%
& 54.5 & 77.6 & 2165 & 81.3 & 86.4 & 93.4\% & 96.1\% \\
\makebox[2.9em][c]{$1$} & 58.2 & 82.0 & 2318 & 84.7 & 87.7 & 98.1\%
& 57.3 & 80.8 & 2290 & 84.1 & 87.3 & 97.1\%
& 54.8 & 77.7 & 2212 & 81.4 & 86.3 & \textbf{94.0\%} & 96.4\% \\
\rowcolor{tableblue!60}
\makebox[2.9em][c]{$2$} & 58.5 & 82.6 & 2332 & 84.7 & 87.8 & \textbf{98.5\%}
& 57.7 & 81.2 & 2297 & 84.0 & 87.3 & \textbf{97.3\%}
& 55.4 & 78.4 & 2178 & 81.3 & 86.3 & \textbf{94.0\%} & \textbf{96.6\%} \\
\makebox[2.9em][c]{$3$} & 58.2 & 82.0 & 2329 & 85.0 & 87.8 & 98.3\%
& 57.5 & 80.7 & 2294 & 84.6 & 87.0 & 97.1\%
& 55.4 & 77.1 & 2161 & 81.3 & 86.4 & 93.6\% & 96.3\% \\
\bottomrule[0.8pt]
\end{tabular}%
}

\medskip
\caption{Ablation of the late endpoint $\ell_d$ on Qwen2.5-VL-7B.
}
\label{tab:ablation-end}
\centering
\resizebox{\linewidth}{!}{%
\begin{tabular}{cccccccccccccccccccc}
\toprule[0.8pt]
& \multicolumn{6}{c}{\textcolor{black!60}{Retain 33.3\%}}
& \multicolumn{6}{c}{\textcolor{black!60}{Retain 22.2\%}}
& \multicolumn{6}{c}{\textcolor{black!60}{Retain 11.1\%}} & \\
\cmidrule(lr){2-7}\cmidrule(lr){8-13}\cmidrule(lr){14-19}
\makebox[2.9em][c]{$\ell_d$} & GQA & MMB & MME & POPE & SQA & RelAcc.
& GQA & MMB & MME & POPE & SQA & RelAcc.
& GQA & MMB & MME & POPE & SQA & RelAcc. & Avg. \\
\midrule[0.4pt]
\makebox[2.9em][c]{$L$} & 58.1 & 82.2 & 2340 & 84.3 & 87.8 & 98.2\%
& 56.8 & 81.8 & 2276 & 82.5 & 87.5 & 96.7\%
& 53.9 & 78.1 & 2143 & 79.7 & 86.3 & 92.8\% & 95.9\% \\
\rowcolor{tableblue!60}
\makebox[2.9em][c]{$L{-}1$} & 58.5 & 82.6 & 2332 & 84.7 & 87.8 & 98.5\%
& 57.7 & 81.2 & 2297 & 84.0 & 87.3 & \textbf{97.3\%}
& 55.4 & 78.4 & 2178 & 81.3 & 86.3 & \textbf{94.0\%} & \textbf{96.6\%} \\
\makebox[2.9em][c]{$L{-}2$} & 58.4 & 82.6 & 2328 & 85.3 & 87.9 & \textbf{98.6\%}
& 57.5 & 81.1 & 2283 & 84.7 & 87.3 & 97.2\%
& 55.3 & 78.2 & 2192 & 81.4 & 85.9 & 93.9\% & \textbf{96.6\%} \\
\makebox[2.9em][c]{$L{-}3$} & 58.4 & 82.0 & 2340 & 85.2 & 87.7 & 98.5\%
& 57.3 & 80.6 & 2295 & 84.3 & 87.2 & 97.1\%
& 55.0 & 77.3 & 2145 & 81.6 & 86.0 & 93.3\% & 96.3\% \\
\bottomrule[0.8pt]
\end{tabular}%
}
\end{table}
\endgroup

\Needspace{8\baselineskip}
\subsection{Number of Groups \texorpdfstring{$K$}{K}}
\label{app:ablation-k}
The group count $K$ controls the granularity of the partition used for budget allocation.
In Table~\ref{tab:ablation-k}, mean RelAcc.\ increases from 94.7\% to 96.6\% as $K$ grows from 8 through 12 and 16 to 20.
Increasing $K$ further to 24 or 28 leaves mean performance at 96.6\% or 96.5\%, respectively, indicating saturation around $K=20$.
This trend, together with the highest RelAcc.\ at 11.1\% retention, supports $K=20$ as the default group count.

\begin{table}[ht]
\centering
\captionsetup{skip=4pt}
\caption{Ablation of the group count $K$ on Qwen2.5-VL.
}
\label{tab:ablation-k}
\setlength{\tabcolsep}{1.6pt}
\renewcommand{\arraystretch}{0.86}
\resizebox{\linewidth}{!}{%
\begin{tabular}{cccccccccccccccccccc}
\toprule[0.8pt]
& \multicolumn{6}{c}{\textcolor{black!60}{Retain 33.3\%}}
& \multicolumn{6}{c}{\textcolor{black!60}{Retain 22.2\%}}
& \multicolumn{6}{c}{\textcolor{black!60}{Retain 11.1\%}} & \\
\cmidrule(lr){2-7}\cmidrule(lr){8-13}\cmidrule(lr){14-19}
\makebox[2.9em][c]{$K$} & GQA & MMB & MME & POPE & SQA & RelAcc.
& GQA & MMB & MME & POPE & SQA & RelAcc.
& GQA & MMB & MME & POPE & SQA & RelAcc. & Avg. \\
\midrule[0.4pt]
\makebox[2.9em][c]{$8$} & 57.6 & 81.2 & 2276 & 83.3 & 87.6 & 97.0\%
& 56.2 & 79.8 & 2263 & 81.8 & 86.7 & 95.5\%
& 53.7 & 76.7 & 2105 & 78.8 & 85.3 & 91.6\% & 94.7\% \\
\makebox[2.9em][c]{$12$} & 57.8 & 80.8 & 2286 & 83.6 & 87.7 & 97.1\%
& 56.2 & 79.7 & 2249 & 82.7 & 86.8 & 95.6\%
& 54.4 & 75.8 & 2131 & 80.0 & 85.6 & 92.2\% & 95.0\% \\
\makebox[2.9em][c]{$16$} & 58.0 & 81.3 & 2302 & 84.4 & 87.8 & 97.7\%
& 56.8 & 79.7 & 2230 & 82.5 & 87.2 & 95.7\%
& 54.6 & 76.6 & 2130 & 80.8 & 85.7 & 92.6\% & 95.3\% \\
\rowcolor{tableblue!60}
\makebox[2.9em][c]{$20$} & 58.5 & 82.6 & 2332 & 84.7 & 87.8 & 98.5\%
& 57.7 & 81.2 & 2297 & 84.0 & 87.3 & \textbf{97.3\%}
& 55.4 & 78.4 & 2178 & 81.3 & 86.3 & \textbf{94.0\%} & \textbf{96.6\%} \\
\makebox[2.9em][c]{$24$} & 58.5 & 81.9 & 2361 & 85.3 & 87.8 & \textbf{98.7\%}
& 57.5 & 81.1 & 2311 & 83.8 & 87.3 & \textbf{97.3\%}
& 55.3 & 78.0 & 2170 & 81.4 & 86.3 & 93.8\% & \textbf{96.6\%} \\
\makebox[2.9em][c]{$28$} & 58.8 & 82.3 & 2331 & 85.0 & 87.9 & 98.6\%
& 57.4 & 80.3 & 2298 & 84.3 & 87.3 & 97.1\%
& 55.5 & 77.9 & 2174 & 81.4 & 86.5 & 93.9\% & 96.5\% \\
\bottomrule[0.8pt]
\end{tabular}%
}
\end{table}

\subsection{Saliency Window Endpoints}
\label{app:saliency-window}
Guided by the layer-wise stage analysis, we compare a small set of representative windows at different depths, including intervals that overlap or cross the sink-dominated stage.
We vary only the saliency window endpoints, with sink filtering enabled and all other components fixed.
An entry $a{\to}b$ measures the displacement between $h^a$ and $h^b$ over $b-a$ layer updates.
Most windows cover five updates; $15{\to}18$ on Qwen2.5-VL and $12{\to}18$ on LLaVA-1.5 additionally probe the sink-dominated stage with different widths.
The default windows follow the stage-based rule in Section~\ref{sec:foreground-window-saliency-estimation}.

Tables~\ref{tab:qwen-window} and~\ref{tab:llava-window} report the results; RelAcc.\ is the mean relative score across GQA, MME, POPE, and SQA.
On both models, the default windows in the late foreground-enhanced stage outperform shallow windows and attain the highest RelAcc.\ among the tested configurations at both retention ratios.
Using the final five encoder updates instead reduces RelAcc.\ by 3.7 and 5.8 percentage points on Qwen2.5-VL and by 2.0 and 2.2 points on LLaVA-1.5 at 22.2\% and 11.1\% retention, respectively.
The gains at the default windows and the decline near the encoder output show the depth dependence of visual saliency estimation and support the window placement for each encoder.

\begin{table}[t]
\centering
\begin{minipage}[t]{0.495\linewidth}
\centering
\captionsetup{hypcap=false,skip=4pt,font=small,justification=raggedright,singlelinecheck=false}
\captionof{table}{Saliency windows on Qwen2.5-VL-7B.}
\label{tab:qwen-window}
\setlength{\tabcolsep}{2.0pt}
\renewcommand{\arraystretch}{0.84}
\scriptsize
\begin{tabular}{lrrrrr}
\toprule[0.8pt]
Endpoints & GQA & MME & POPE & SQA & RelAcc. \\
\midrule[0.4pt]
\multicolumn{6}{l}{\textcolor{black!60}{Uncompressed baseline (100\%)}} \\
Vanilla & 60.9 & 2310 & 86.3 & 88.9 & 100\% \\
\specialrule{0.3pt}{3pt}{3pt}
\multicolumn{6}{l}{\textcolor{black!60}{Token retention: 22.2\%}} \\
$0{\to}5$ & 57.2 & 2269 & 84.0 & 87.3 & 96.9\% \\
$2{\to}7$ & 57.2 & 2273 & 84.2 & 87.4 & 97.1\% \\
$7{\to}12$ & 57.1 & 2268 & 84.0 & 87.3 & 96.9\% \\
$12{\to}17$ & 57.7 & 2269 & 83.9 & 87.3 & 97.1\% \\
$15{\to}18$ & 57.5 & 2290 & 84.0 & 87.1 & 97.2\% \\
$15{\to}20$ & 57.7 & 2291 & 84.0 & 87.2 & 97.3\% \\
\rowcolor{tableblue!60}
Default $19{\to}24$ & 57.7 & 2297 & 84.0 & 87.3 & \textbf{97.4\%} \\
$24{\to}29$ & 57.2 & 2272 & 84.0 & 87.3 & 97.0\% \\
$27{\to}32$ & 54.1 & 2245 & 79.6 & 86.0 & 93.7\% \\
\specialrule{0.3pt}{3pt}{3pt}
\multicolumn{6}{l}{\textcolor{black!60}{Token retention: 11.1\%}} \\
$0{\to}5$ & 54.3 & 2134 & 80.9 & 86.3 & 93.1\% \\
$2{\to}7$ & 54.7 & 2159 & 80.9 & 86.6 & 93.6\% \\
$7{\to}12$ & 54.6 & 2172 & 81.1 & 86.7 & 93.8\% \\
$12{\to}17$ & 54.9 & 2138 & 81.2 & 86.3 & 93.5\% \\
$15{\to}18$ & 55.0 & 2123 & 81.5 & 86.2 & 93.4\% \\
$15{\to}20$ & 55.3 & 2150 & 81.5 & 86.3 & 93.8\% \\
\rowcolor{tableblue!60}
Default $19{\to}24$ & 55.4 & 2178 & 81.3 & 86.3 & \textbf{94.1\%} \\
$24{\to}29$ & 54.5 & 2207 & 81.3 & 86.1 & 94.0\% \\
$27{\to}32$ & 51.3 & 2018 & 74.2 & 85.1 & 88.3\% \\
\bottomrule[0.8pt]
\end{tabular}
\end{minipage}%
\hfill
\begin{minipage}[t]{0.485\linewidth}
\centering
\captionsetup{hypcap=false,skip=4pt,font=small,justification=raggedright,singlelinecheck=false}
\captionof{table}{Saliency windows on LLaVA-1.5-7B.}
\label{tab:llava-window}
\setlength{\tabcolsep}{2.0pt}
\renewcommand{\arraystretch}{0.84}
\scriptsize
\begin{tabular}{lrrrrr}
\toprule[0.8pt]
Endpoints & GQA & MME & POPE & SQA & RelAcc. \\
\midrule[0.4pt]
\multicolumn{6}{l}{\textcolor{black!60}{Uncompressed baseline (100\%)}} \\
Vanilla & 61.9 & 1862 & 85.9 & 69.5 & 100\% \\
\specialrule{0.3pt}{3pt}{3pt}
\multicolumn{6}{l}{\textcolor{black!60}{Token retention: 22.2\%}} \\
$0{\to}5$ & 58.4 & 1691 & 86.5 & 68.1 & 96.0\% \\
$2{\to}7$ & 58.4 & 1690 & 86.3 & 68.5 & 96.0\% \\
$7{\to}12$ & 58.5 & 1744 & 86.0 & 68.3 & 96.6\% \\
$12{\to}17$ & 59.2 & 1762 & 86.8 & 68.5 & 97.5\% \\
$12{\to}18$ & 59.3 & 1752 & 86.7 & 68.5 & 97.3\% \\
\rowcolor{tableblue!60}
Default $14{\to}19$ & 58.9 & 1778 & 86.7 & 68.7 & \textbf{97.6\%} \\
$19{\to}24$ & 57.5 & 1710 & 85.7 & 68.1 & 95.6\% \\
\specialrule{0.3pt}{3pt}{3pt}
\multicolumn{6}{l}{\textcolor{black!60}{Token retention: 11.1\%}} \\
$0{\to}5$ & 56.6 & 1687 & 85.6 & 67.9 & 94.8\% \\
$2{\to}7$ & 56.6 & 1655 & 85.0 & 68.8 & 94.6\% \\
$7{\to}12$ & 56.5 & 1647 & 85.8 & 68.0 & 94.4\% \\
$12{\to}17$ & 57.7 & 1689 & 85.5 & 68.6 & 95.5\% \\
$12{\to}18$ & 57.6 & 1687 & 85.4 & 68.0 & 95.2\% \\
\rowcolor{tableblue!60}
Default $14{\to}19$ & 57.7 & 1708 & 85.6 & 68.4 & \textbf{95.8\%} \\
$19{\to}24$ & 54.9 & 1658 & 84.5 & 68.2 & 93.6\% \\
\bottomrule[0.8pt]
\end{tabular}
\end{minipage}
\end{table}

\subsection{Saliency Window Width}
\label{app:saliency-window-length}
To test sensitivity to the empirical five-layer width, we fix the start at Layer 19 on Qwen2.5-VL and vary the window width $w=\ell_e-\ell_s$, keeping all other components unchanged.
The default $w=5$ corresponds to $19{\to}24$; $w=3$ and $w=7$ give $19{\to}22$ and $19{\to}26$.
Table~\ref{tab:qwen-window-length} reports results at 22.2\% and 11.1\% retention.
The default width attains the highest RelAcc.\ at 22.2\% retention.
At 11.1\% retention, $w=3$ is 0.2 points higher than the default, while $w=7$ is 0.4 points lower.
Across $w\in\{3,5,7\}$, the RelAcc.\ range is 0.4 and 0.6 percentage points at the two retention ratios, respectively.
These small variations indicate limited sensitivity to window width over the tested range.
We use $w=5$ as a shared empirical default throughout our experiments.
\begin{table}[ht]
\centering
\captionsetup{skip=4pt}
\caption{Saliency window width ablation on Qwen2.5-VL-7B with the start fixed at Layer $19$ and all other components unchanged.
An interval $a{\to}b$ covers updates from Layer $a$ through Layer $b-1$, so varying $b$ changes the width while preserving the starting layer.}
\label{tab:qwen-window-length}
\setlength{\tabcolsep}{2.6pt}
\renewcommand{\arraystretch}{0.9}
\scriptsize
\begin{tabular}{lrrrrr}
\toprule[0.8pt]
Window & GQA & MME & POPE & SQA & RelAcc. \\
\midrule[0.4pt]
\multicolumn{6}{l}{\textcolor{black!60}{Uncompressed baseline (100\%)}} \\
Vanilla & 60.9 & 2310 & 86.3 & 88.9 & 100\% \\
\specialrule{0.3pt}{3pt}{3pt}
\multicolumn{6}{l}{\textcolor{black!60}{Token retention: 22.2\%}} \\
$19{\to}22$ ($w{=}3$) & 57.8 & 2252 & 84.2 & 87.1 & 97.0\% \\
\rowcolor{tableblue!60}
$19{\to}24$ ($w{=}5$) & 57.7 & 2297 & 84.0 & 87.3 & \textbf{97.4\%} \\
$19{\to}26$ ($w{=}7$) & 57.6 & 2273 & 84.0 & 87.4 & 97.2\% \\
\specialrule{0.3pt}{3pt}{3pt}
\multicolumn{6}{l}{\textcolor{black!60}{Token retention: 11.1\%}} \\
$19{\to}22$ ($w{=}3$) & 55.6 & 2182 & 81.3 & 86.6 & \textbf{94.3\%} \\
\rowcolor{tableblue!60}
$19{\to}24$ ($w{=}5$) & 55.4 & 2178 & 81.3 & 86.3 & 94.1\% \\
$19{\to}26$ ($w{=}7$) & 55.2 & 2147 & 81.0 & 86.4 & 93.7\% \\
\bottomrule[0.8pt]
\end{tabular}
\end{table}

\Needspace{12\baselineskip}
\subsection{Signal Assignment}
\label{app:signal-assignment}
The full method uses the query-weighted saliency score $s_i=\alpha_i u_i$ for both budget allocation across groups and token ranking within each group.
Table~\ref{tab:ablation-selection} compares two variants: (i) the query relevance score $\alpha_i$ for allocation and the visual saliency score $u_i$ for ranking, and (ii) $u_i$ for allocation and $\alpha_i$ for ranking.
Grouping and the remaining pipeline are held fixed.
Using query relevance for allocation and visual saliency for ranking exceeds the reverse assignment by 2.5 and 2.7 percentage points and remains within 0.1 point of the full method.
These results support using query relevance for budget allocation across groups and visual saliency for token selection within each group.

\begin{table}[ht]
\centering
\captionsetup{skip=4pt}
\caption{Ablation of signal assignment on Qwen2.5-VL-7B.
Allocation and ranking refer to budget allocation across groups and token selection within each group, respectively.}
\label{tab:ablation-selection}
\setlength{\tabcolsep}{2.6pt}
\renewcommand{\arraystretch}{0.9}
\scriptsize
\begin{tabular}{lrrrrr}
\toprule[0.8pt]
Method & GQA & MME & POPE & SQA & RelAcc. \\
\midrule[0.4pt]
\multicolumn{6}{l}{\textcolor{black!60}{Token retention: 22.2\%}} \\
$\alpha$ Alloc., $u$ Rank & 57.4 & 2307 & 84.2 & 87.1 & \textbf{97.4\%} \\
$u$ Alloc., $\alpha$ Rank & 55.0 & 2264 & 81.1 & 86.6 & 94.9\% \\
\rowcolor{tableblue!60}
Ours & 57.7 & 2297 & 84.0 & 87.3 & \textbf{97.4\%} \\
\specialrule{0.3pt}{3pt}{3pt}
\multicolumn{6}{l}{\textcolor{black!60}{Token retention: 11.1\%}} \\
$\alpha$ Alloc., $u$ Rank & 55.0 & 2166 & 81.5 & 86.5 & 94.0\% \\
$u$ Alloc., $\alpha$ Rank & 52.7 & 2113 & 78.0 & 86.1 & 91.3\% \\
\rowcolor{tableblue!60}
Ours & 55.4 & 2178 & 81.3 & 86.3 & \textbf{94.1\%} \\
\bottomrule[0.8pt]
\end{tabular}
\end{table}

\clearpage
\section{Details of Evaluation Benchmarks}
\label{app:benchmarks}
We summarize the tasks covered by the evaluation benchmarks and define their notation in the result tables.

\noindent\textbf{GQA}~\citep{hudson2019gqa}.
GQA evaluates compositional visual reasoning through questions about objects, attributes, and spatial relations.
Its questions are constructed from Visual Genome scene graphs~\citep{krishna2017visual}.

\noindent\textbf{MMBench}~\citep{liu2024mmbench}.
MMBench evaluates visual perception and reasoning through questions with multiple answer choices in English and Chinese, covering 20 ability dimensions.
We use the English split and denote it as MMB in the tables.

\noindent\textbf{MME}~\citep{NEURIPS2025_d79a27cf}.
MME evaluates MLLMs across 14 perception and cognition tasks, including object recognition, spatial reasoning, text recognition, and knowledge reasoning.
In our tables, MME refers to the full perception and cognition suite.

\noindent\textbf{POPE}~\citep{li2023evaluating}.
POPE assesses object hallucination through questions about whether specified objects are present in an image.
We use the configuration based on MS COCO images~\citep{lin2014microsoft}.
We report the average F1 score over the adversarial, random, and popular settings.

\noindent\textbf{ScienceQA}~\citep{lu2022learn}.
ScienceQA evaluates reasoning over questions spanning natural science, social science, and language science.
Questions may include image or text context.
We denote the benchmark as SQA in the tables.

\noindent\textbf{VQAv2}~\citep{goyal2017making}.
VQAv2 is a visual question answering benchmark containing 204,721 images from MS COCO~\citep{lin2014microsoft} and 1,105,904 questions.
It includes complementary image pairs with different answers to reduce reliance on language priors.
Each question is associated with ten human answers.
We denote the benchmark as VQA$^{\text{v2}}$ in the tables.

\noindent\textbf{TextVQA}~\citep{singh2019towards}.
TextVQA evaluates reading and reasoning over text in natural images.
It contains 45,336 questions associated with 28,408 images from the Open Images dataset~\citep{krasin2017openimages}.
We denote the benchmark as VQA$^{\text{Text}}$ in the tables.

\noindent\textbf{MMMU}~\citep{yue2024mmmu}.
MMMU evaluates multimodal knowledge and reasoning at the college level across disciplines such as art, business, science, and engineering.

\noindent\textbf{SEED-Bench}~\citep{li2023seed}.
SEED-Bench evaluates visual understanding through questions with multiple answer choices and human annotations.
It includes tasks based on static images and videos; SEED$^{\text{I}}$ denotes the image split used in all our result tables.

\noindent\textbf{VizWiz}~\citep{gurari2018vizwiz}.
VizWiz contains photographs and questions collected from visually impaired users.
It includes images of varying quality and questions that cannot always be answered from the available visual content.

\noindent\textbf{HR-Bench 8K}~\citep{wang2025divide}.
HR-Bench 8K is the subset of HR-Bench constructed from 8K images in DIV8K~\citep{gu2019div8k}.
It evaluates the ability to recover fine visual details from images at 8K resolution.
We denote the benchmark as HRB$^{\text{8K}}$ in the tables.

\clearpage
\section{Qualitative Analysis of Retained Tokens}
\label{app:viz}

We examine the spatial distribution of retained tokens to better understand the behavior of \ours{} under compression.
We first analyze two cases where pruning changes a correct answer into an incorrect one, then compare token selection across \ours{}, VisionZip, and PruneSID.

\subsection{Failure Cases}
\label{app:failure-cases}
\begin{figure}[htbp]
\centering
\includegraphics[width=\linewidth]{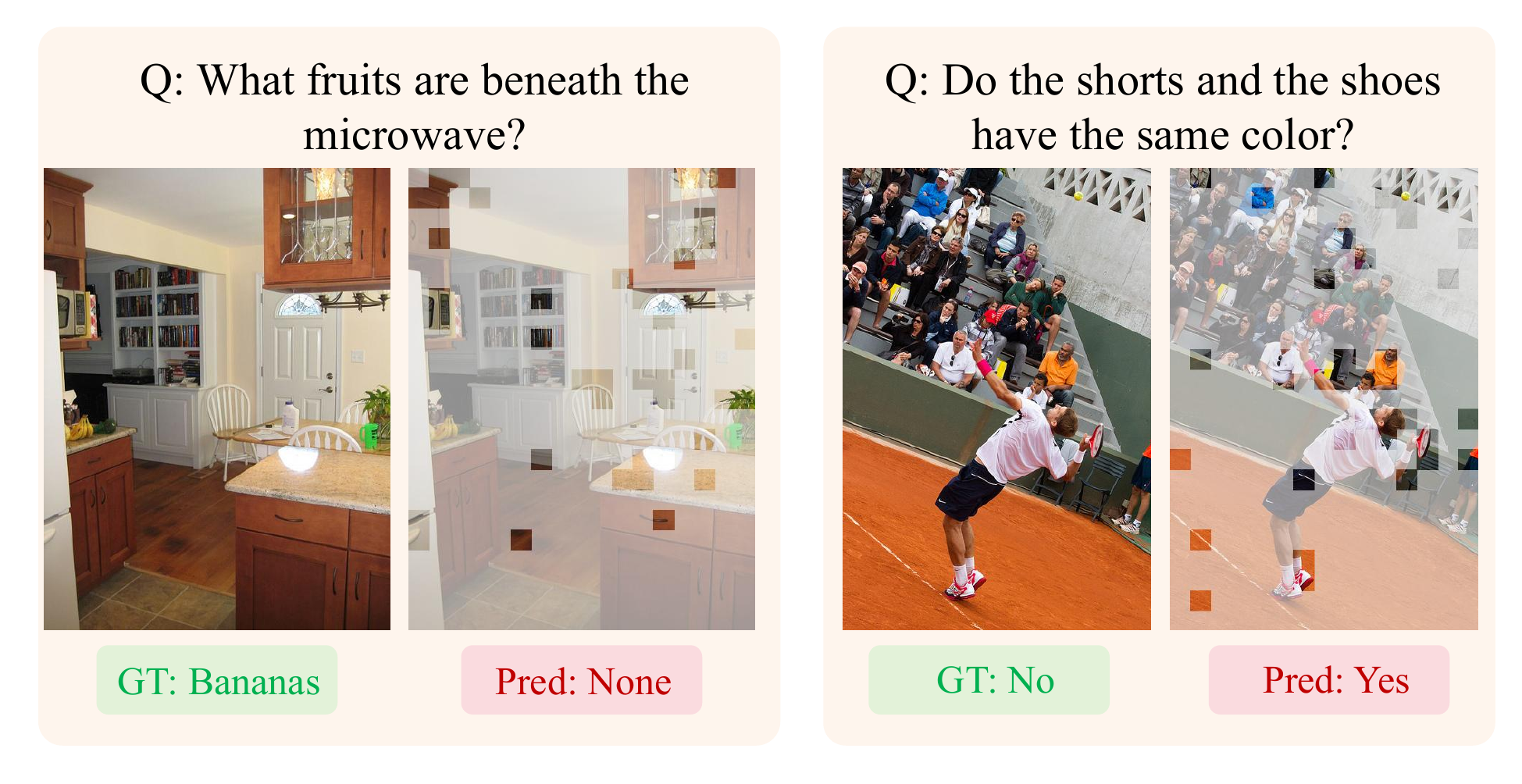}
\caption{Failure cases of \ours{} on Qwen2.5-VL-7B at 11.1\% target token retention.
Each case pairs the original image (left) with the retained-token visualization (right).
GT denotes the ground-truth answer and Pred denotes the answer after pruning; the uncompressed model answers both questions correctly.
Discarded patches are covered by a translucent gray mask.}
\label{fig:app-failure-cases}
\end{figure}

Figure~\ref{fig:app-failure-cases} shows two GQA examples on Qwen2.5-VL-7B at a target token retention ratio of 11.1\%.
The uncompressed model answers both questions correctly, while \ours{} fails after pruning.
In the kitchen example, the bananas occupy a small region beneath the microwave and receive limited coverage from the retained tokens.
The pruned model answers ``none'' instead of ``bananas,'' illustrating a failure to identify a small object required by the query.

In the tennis example, the pruned model incorrectly answers that the shorts and shoes have the same color.
Although parts of both objects remain represented by retained patches, the model fails to distinguish their colors.
These cases highlight the difficulty of preserving sufficient evidence for small-object recognition and attribute comparison within a limited token budget, even when selection incorporates query relevance.

\clearpage
\subsection{Comparison of Retained Tokens}
\label{app:retained-token-comparison}
\begin{figure}[!b]
\centering
\includegraphics[width=\linewidth,height=0.78\textheight,keepaspectratio]{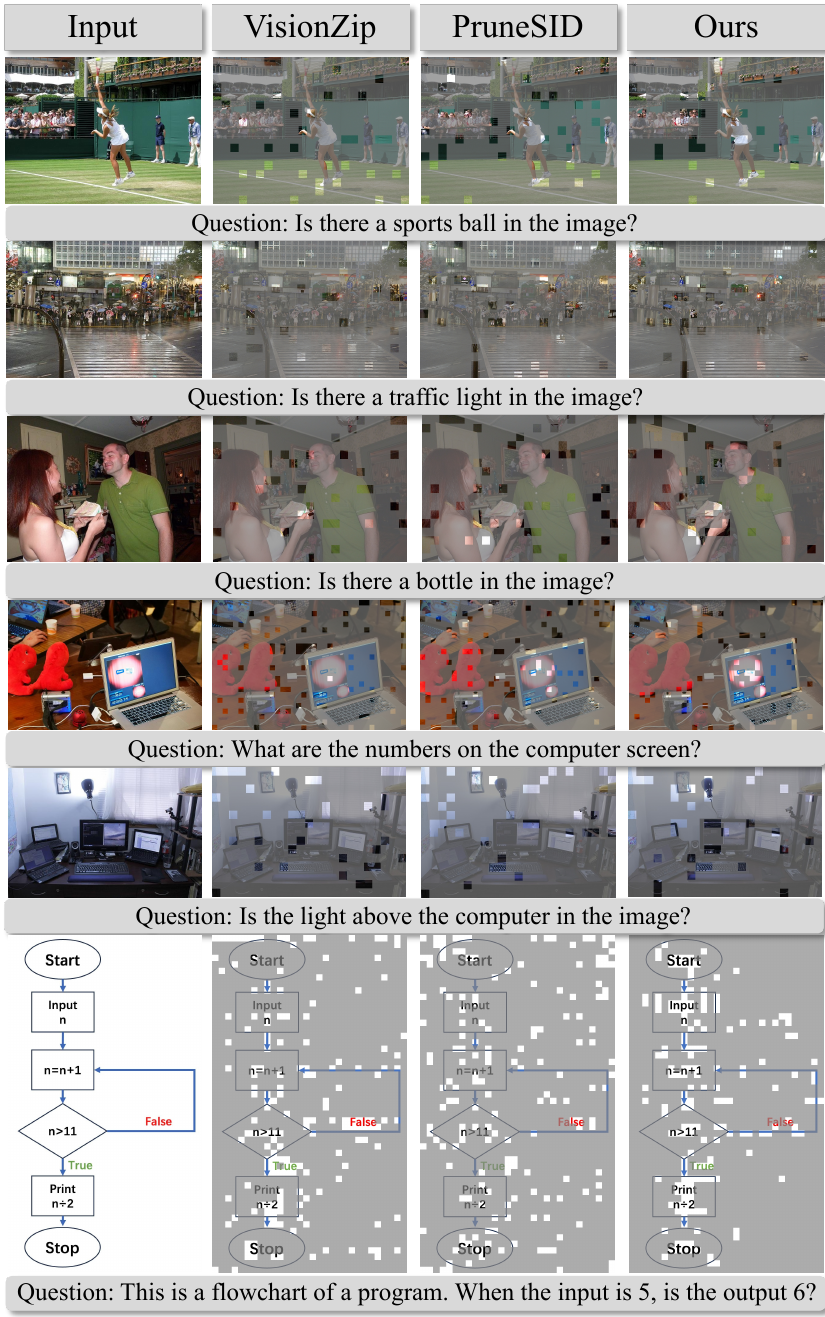}
\caption{Qualitative comparison at a retention ratio of 11.1\%.
Columns show the original image and the tokens retained by VisionZip, PruneSID, and \ours{}, with the corresponding query below each example.
Discarded patches are covered by a translucent gray mask.}
\label{fig:app-viz}
\end{figure}

Figure~\ref{fig:app-viz} compares the tokens retained by VisionZip, PruneSID, and \ours{} for the same images and queries.
The examples cover object presence, text reading, spatial relations, and flowchart reasoning at 11.1\% token retention.
For the laptop and flowchart queries, \ours{} retains patches over portions of the screen content and program structure.
For the spatial query, retained patches cover both the lamp and computer regions.
These examples illustrate the coverage of queried content across natural images and diagrams, complementing the quantitative benchmark results.

\end{document}